\documentclass[runningheads]{llncs}
\pdfoutput=1
 
\usepackage{eccv}

\usepackage{eccvabbrv}

\usepackage{graphicx}
\usepackage[utf8]{inputenc} %
\usepackage[T1]{fontenc}    %
\usepackage{hyperref}       %
\usepackage{url}            %
\usepackage{booktabs}       %
\usepackage{amsfonts}       %
\usepackage{nicefrac}       %
\usepackage{xcolor}         %
\usepackage{amsmath}
\usepackage{wrapfig}
\usepackage{float}
\usepackage{microtype}      %
\usepackage{multirow}
\usepackage{wrapfig}

\newcommand{\shiry}[1]{{\color{orange}[shiry: #1]}}

\newcommand{\carl}[1]{{\color{blue}[carl: #1]}}
\newcommand{\neerja}[1]{{\color{magenta}[neerja: #1]}}

\renewcommand{\neerja}[1]{}
\renewcommand{\carl}[1]{}
\renewcommand{\shiry}[1]{}
\newcommand{\starinst}{\textsuperscript{*}}
\usepackage[accsupp]{axessibility}  %

\usepackage{hyperref}

\usepackage{orcidlink}

\begin{document}

\title{Forecasting %
Motion in the Wild}

\titlerunning{Forecasting Motion in the Wild}

\author{Neerja Thakkar\inst{1,2} \and
Shiry Ginosar\starinst\,\inst{3} \and
Jacob Walker\inst{2} \and 
Jitendra Malik\starinst\, \inst{1} \and
Joao Carreira\starinst\, \inst{2} \and 
Carl Doersch\starinst\,\inst{2}
}

\authorrunning{N.~Thakkar et al.}

\institute{UC Berkeley \and
Google DeepMind \and 
Toyota Technological Institute at Chicago}

\maketitle

\renewcommand{\thefootnote}{\fnsymbol{footnote}}
\footnotetext[1]{Equal advising contribution.}
\renewcommand{\thefootnote}{\arabic{footnote}}

\begin{figure*}[t]
  \centering
    \includegraphics[width=0.99\linewidth]{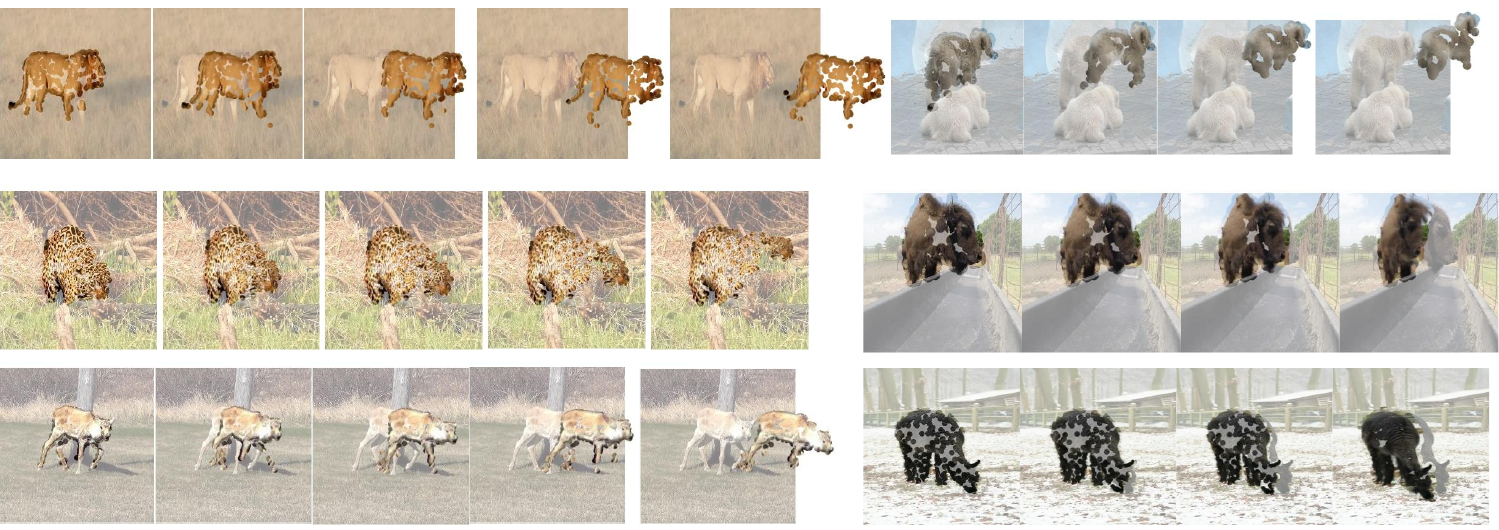}
  \caption{\textbf{Dense point trajectories act as visual tokens for behavior, enabling scalable prediction of complex motion across diverse species.} Our method takes as input a single RGB image, a history of motion, and an optional high-level motion vector, and forecasts future animal motion in the form of point trajectories. 
  For each predicted point trajectory, we translate a small circular patch of the input image along the motion trajectory and superimpose it on the input image (\textbf{no pixels are generated!}).  Leftmost shows the start locations on the input frame; the rest is forecast by our model.
  Our method is capable of forecasting many different animal species and behaviors, even long-tail ones---the polar bear on the top right is only present in $0.31\%$ of the training data, the caribou on the bottom left in $0.025\%$, and the alpaca on the bottom right in $0.50\%$. See more results at https://motion-forecasting.github.io/.}
  \label{fig:teaser}
  \vspace{-0.9em}
\end{figure*}

\begin{abstract}
Visual intelligence requires anticipating the future behavior of agents, yet vision systems lack a general representation for motion and behavior. We propose dense point trajectories as visual tokens for behavior, a structured mid-level representation that disentangles motion from appearance and generalizes across diverse non-rigid agents, such as animals in-the-wild. Building on this abstraction, we design a diffusion transformer that models unordered sets of trajectories and explicitly reasons about occlusion, enabling coherent forecasts of complex motion patterns. To evaluate at scale, we curate 300 hours of unconstrained animal motion from video through robust shot detection and camera-motion compensation. Experiments show that forecasting trajectory tokens achieves category-agnostic, data-efficient prediction, outperforms state-of-the-art baselines, and generalizes to rare species and morphologies, providing a foundation for predictive visual intelligence in the wild.

\end{abstract}

\section{Introduction}
\label{sec:intro}

Predicting the future motion of objects and agents is a fundamental capability of visual intelligence. In dynamic environments, agents, from animals in the wild to humans in social settings, must anticipate the behavior of others in order to act effectively or survive. Despite major advances in visual recognition and generation, predicting behavior remains one of the least understood capabilities of modern vision systems.

A key reason for this gap is the lack of an appropriate representation for behavior. In language, prediction is enabled by discrete tokens that structure the modeling problem. Vision systems lack an analogous token for motion and behavior. In this work, we show that dense point trajectories can serve as such tokens, enabling scalable prediction of behavior across diverse agents.
To understand why such a representation is needed, consider the limitations of existing approaches.
Forecasting directly in pixel space is universal but poorly structured: while recent video diffusion models can generate realistic short clips, forecasting behavior directly in pixel space entangles appearance, lighting, and camera motion with object dynamics, making the learning problem unnecessarily complex and data inefficient. At the opposite extreme, parameterized 3D models provide compact and physically valid representations for forecasting, but rely on strong object-specific priors and therefore apply only to a small number of carefully modeled categories, such as humans~\cite{loper2023smpl} and \carl{a handful of } animals~\cite{zuffi20173d,zuffi2018lions,zuffi2019three,ruegg2023bite,Zuffi_2024_CVPR,wu2023magicpony,sun2024ponymation}. Even in these settings they often miss fine-grained deformation and shape variation. Without an intermediate representation that captures motion structure while remaining general, scalable behavior prediction remains difficult. We therefore seek a representation that introduces structure without sacrificing generality.

We therefore propose dense point trajectories as visual tokens for behavior, providing a structured representation for forecasting motion across diverse agents. While sparse points carry little semantic meaning when static, their motion reveals rich information about 3D structure and intent, as demonstrated in Johansson's classical biological motion studies~\cite{johansson1973visual} and subsequent work~\cite{kozlowski1977recognizing,cutting1977recognizing,grossman2000brain,fox1982perception,atkinson2004emotion}. Representing behavior as evolving 2D point tracks focuses prediction directly on motion dynamics while remaining agnostic to appearance and scene variation. This formulation is significantly more data efficient than forecasting pixels directly~\cite{bharadhwaj2024track2act,bharadhwaj2025gen2act} and naturally applies to arbitrary non-rigid agents without requiring category-specific models. Point trajectories therefore occupy a principled middle ground between raw pixels and full 3D parameterizations: structured enough to constrain prediction, yet general enough to scale across species, morphologies, and environments.

Building on this abstraction, we introduce a diffusion transformer that forecasts behavior from short motion histories. Unlike prior trajectory-based approaches designed for robotics or rigid scenes~\cite{bharadhwaj2024track2act,wen2023anypoint,Chen_2025_ICCV}, our formulation models motion for non-rigid agents in the wild. The model predicts future behavior as an unordered set of point trajectories (Fig.~\ref{fig:teaser}), treating each trajectory as a token augmented with local visual context from DINOv3 features. The architecture jointly models trajectories while explicitly reasoning about occlusion and visibility, enabling coherent predictions of complex non-rigid motion. Our model learns diverse motion patterns including gait, cyclical, and linear behaviors, and forecasts future motion across a wide range of species, outperforming state-of-the-art baselines. Training on the broad diversity of motion found in nature further enables generalization to previously unseen categories and morphologies of animate agents.

To study long-tailed motion at scale, we focus on unconstrained video of animals in the wild. Animals provide a particularly challenging testbed for motion prediction: they exhibit highly diverse morphologies and behavior patterns, and data for many species is inherently sparse. A representation that succeeds in this regime must generalize across categories without relying on category-specific models. We develop a large-scale pipeline for isolating animal motion from raw video, including robust shot detection and camera-motion compensation, and curate 300 hours of annotated footage for behavior forecasting which we release with this paper. Using this in-the-wild data, we demonstrate that our approach operates on tracks extracted from unconstrained video and is robust to the noise and partial observability inherent in real-world tracking. This dataset reveals previously unreported statistical structure in animal motion, and provides a foundation for studying predictive visual intelligence in natural environments.

\carl{this paragraph seems redundant?} Our results show that forecasting point trajectories enables structured, category-agnostic prediction of complex real-world behavior. These results suggest that trajectory tokens provide a scalable foundation for predictive visual intelligence.

Our contributions are:
\begin{enumerate}
\item \textbf{Point trajectories as visual tokens for behavior forecasting.}
We introduce dense point tracks as a compact mid-level representation for modeling long-tailed natural-world behavior that disentangles motion from appearance and generalizes beyond category-specific 3D models.

\item \textbf{A diffusion transformer for trajectory forecasting.}
We design a DiT-based architecture that treats trajectories as tokens and predicts diverse futures of non-rigid behavior from short histories while explicitly reasoning about occlusion in unordered track sets.

\item \textbf{MammalMotion, a large-scale dataset of animal motion.}
We develop a robust pipeline for isolating animal motion in unconstrained video and release 300 hours of annotated footage for behavior forecasting in the wild.

\end{enumerate}

\section{Related Work}
\label{sec:related_work}

\noindent
\textbf{Pixel Forecasting.} When it comes to forecasting visual information, pixels have been the natural choice for several years. Early approaches predicted future pixels deterministically, as a regression problem~\cite{ranzato2014video, srivastava2015unsupervised, 9294028}, which is exceedingly challenging, since the problem is ambiguous, and leads to blurry predictions. 

While GANs~\cite{clark2019adversarial,tulyakov2018mocogan,wang2020imaginator} and variational models ~\cite{lee2018stochastic} were once promising, many modern approaches use diffusion models~\cite{ho2020denoising} which produce sharp videos~\cite{gu2023seer,xing2024aid,hoppe2022diffusion,ye2024stdiff,yang2023video,gupta2024walt} -- and have brought on a creative video revolution~\cite{openai-2024, google_veo3_2025}.
However, training models directly on video is expensive and data-inefficient and models still struggle with hallucinations and basic physical interactions~\cite{chefer2025, bansalvideophy,kang2025far}. 

\medskip
\noindent
\textbf{Point Track Forecasting.} Several works have pushed the frontier in high-quality point-tracking~\cite{doersch2023tapir,doersch2024bootstap,karaev2024cotracker,karaev2025cotracker3,zholus2025tapnext}, with broad applications across different computer vision tasks. When it comes to forecasting point tracks, the most significant advancements have come from the robotics domain. Any-point Trajectory Modeling~\cite{wen2023anypoint} introduced the paradigm of first training a regression model to predict point tracks from an image and language instruction, and learning a robot policy on top of the track prediction model. Several approaches have followed in this direction~\cite{bharadhwaj2024track2act,flip,xu2024flow,yuan2024general,Chen_2025_ICCV,yang2025tra,niu2025pre}. These works have explored different architectures for forecasting point tracks such as conditional diffusion transformers~\cite{bharadhwaj2024track2act,Chen_2025_ICCV} and latent diffusion models~\cite{xu2024flow}, all with the end-goal of learning good robotic manipulation policies. Similarly,~\cite{walker2025} applies DiTs to forecast frozen video encodings along with future decoded point tracks. We draw inspiration from these conditional DiT architectures but focus on a different application, forecasting motion in the complex domain of in-the-wild animal data.  

Most recently, \cite{boduljak2025happens} showed that point-track forecasting outperforms pixel generation for simple Kubric~\cite{greff2022kubric} object motions. Our work provides further evidence that point tracks can be a more data-efficient representation for motion, by expanding their scope to more challenging and non-rigid domain of in-the-wild animal data.

\medskip
\noindent
\textbf{Behavioral Forecasting in Computer Vision.} 

Beyond pixels and tracks, there has also been work focusing on forecasting the behavior of intelligent entities as well as their interactions. For example, human trajectory prediction has a long history with a variety of approaches~\cite{kitani2012activity, rudenko2020human}. For direct trajectory prediction, these range from RNN based approaches~\cite{salzmann2020trajectron++, vemula2018social} to VAEs~\cite{mangalam2020not} and GANS ~\cite{gupta2018social}, leveraging  generative modeling of future human trajectories. Many recent papers also focus on utilizing scene context~\cite{salzmann2023robots, thakkar2024adaptive} for human trajectory forecasting. Behavioral forecasting has also been extensively explored in the context of autonomous driving~\cite{seff2023motionlm, Li_2025_ICCV, moon2024visiontrap, lee2017desire}. Relatively few vision papers have focused on forecasting animal motion. QuadForecaster~\cite{noronhaquadforecaster} predicted the poses of animals in constrained contexts while~\cite{Liu_2021_ICCV} demonstrated a proof of concept of their approach on fish and mice. In contrast, our approach leverages large and diverse datasets and forecasts animal motion on a general level.

\medskip
\noindent
\textbf{Animal Pose, Motion and Behavior.} Ethology, the study of animal behavior, has a long history~\cite{tinbergen1963aims, lorenz1938taxis, lorenz1935kumpan, von1953dancing}. Recent advances in computing and machine learning
show promise in aiding discoveries -- e.g. the emerging field of Computational Ethology~\cite{anderson2014toward} where computer vision and automated motion analysis plays a major role.  For example, work such as DeepLabCut~\cite{mathis2018deeplabcut, nath2019using, lauer2022multi} and SLEAP~\cite{pereira2022sleap} have accelerated annotating poses of animals in video. Video analysis has aided the ethology of a wide range of animals, from jumping plant lice~\cite{polajnar2024wing}, mice~\cite{ye2024superanimal} and even zebrafish larvae~\cite{scholz2025plug}. However, as~\cite{anderson2014toward} notes, for most of these approaches, humans still need to manually annotate behaviors in training data -- which can be subjective due to varied spatial and temporal scales, and limited by human perception and difficulties in  discovering new behaviors. Our work leverages massive datasets of unlabeled videos and is a step towards automatic motion understanding of animals.

There has also been a line of work on reconstructing animal pose in 3D~\cite{zuffi20173d, zuffi2018lions}, moving towards accurate reconstructions of individual species~\cite{zuffi2019three,ruegg2023bite,Zuffi_2024_CVPR,wu2023magicpony}, or creating species-specific models to generate 3D animal motion~\cite{sun2024ponymation}. These works provide insights into individual animal species, but our work focuses on developing an approach that is data-efficient and can generalize to many species including long tail ones.

\section{Forecasting Point Trajectories with a Diffusion Model}

\begin{figure*}[t]
  \centering
    \includegraphics[width=0.9\linewidth]{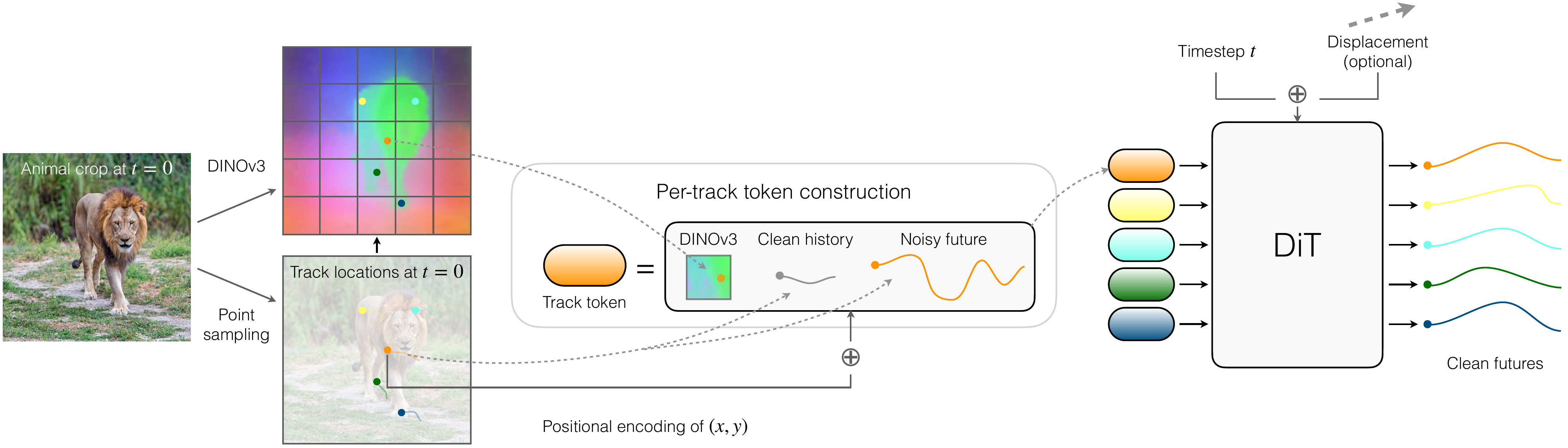}
  \caption{\textbf{Architecture.} Given an input frame and (noisy) tracks, we construct a single token for every track, which includes a DINO feature at the start location, the motion history, and the noisy track values, both with occlusion indicators.  After projection, we add a position encoding for the initial point location.  Tokens are stacked and fed to a transformer (DiT) to predict clean tracks (right).}
  \label{fig:architecture}
  \vspace{-0.9em}
\end{figure*}

We present a diffusion-based approach for generating animal motion as a sequence of point tracks. Unlike video generation models that predict RGB pixels, our method operates directly on point trajectories.
Given a single observation frame and optional conditioning information like motion history or desired velocity, our model generates plausible future trajectories.

\subsection{Problem Formulation}

We represent the motion of a single animal as a set of $N$ point tracks, where each track describes the 2D trajectory of a single surface point over a time horizon of $T$ timesteps. Formally, we aim to predict a set of tracks $\mathbf{X} \in \mathbb{R}^{T \times N \times 2}$. Each point track $\mathbf{x}_n = [(x_n^1, y_n^1), (x_n^2, y_n^2), \ldots, (x_n^{T}, y_n^{T})]$, consists of a sequence of normalized coordinates $(x_n^t, y_n^t)$ where $t$ indexes time. Points may become occluded, in which case we assume the location is unknown: we represent the occlusion state as $\mathbf{O} \in \mathbb{R}^{T \times N}$, where $\mathbf{O}_{n}^t \in [0, 1]$ indicates that it the $n$'th point is visible ($1$) or occluded ($0$) at time $t$.

Our forecasting model learns a conditional generative distribution:

$$p(\mathbf{X}_{T_c+1:T}, \mathbf{O}_{T_c+1:T} | \mathbf{I}, \mathbf{X}_{1:T_c}, \mathbf{O}_{1:T_c}, \mathbf{d})$$

Where $\mathbf{I}$ is the first frame, $\mathbf{X}_{1:T_c}$ and $\mathbf{O}_{1:T_c}$ are the observed conditioning motion history tracks and occlusion states over the first $T_c$ timesteps, and a single optional 2D displacement vector $\mathbf{d} \in \mathbb{R}^2$ describing the average motion of tracks from the last frame: $\mathbf{d} = \sum_{n=1}^N \mathbf{O}_{n}^T[(x_n^{T}, y_n^{T}) - (x_n^1, y_n^1)] / \sum_{n=1}^N \mathbf{O}_{n}^T$.
The model generates future trajectories $\mathbf{X}_{T_c+1:T}$ and occlusion states $\mathbf{O}_{T_c+1:T}$ conditioned on this observed history and the optional conditioning.  Because the main challenge is to predict the track positions $\mathbf{X}_{T_c+1:T}$, which are a high-dimensional and continuous value, we draw inspiration from prior work~\cite{bharadhwaj2024track2act} and model distribution with a diffusion process.

\medskip
\noindent
\textbf{Parameterization of the diffusion target.}  Diffusion involves adding Gaussian noise to the inputs (tracks and occlusions) and training a network to denoise them.  While we could directly denoise $\mathbf{X}$ and $\mathbf{O}$, there are two problems.  First, $\mathbf{X}$ has missing values for occluded points (prior work, e.g.~\cite{wen2023anypoint}, assumes there are no missing points, which is untenable for longer horizons).  Second, $\mathbf{X}$ values are extremely correlated, and most of the variance is due to the initial point that is tracked rather than due to the motion itself.  We therefore reparameterize the tracks to improve training dynamics.  Specifically, we construct the diffusion target $\mathbf{Z}_0^{\text{diff}}=\{\gamma\mathbf{V},\beta\mathbf{O}\}$ where the $n$'th row of $\mathbf{V}\in \mathbb{R}^{N\times T\times 2}$ is $[(\dot{x}_n^1, \dot{y}_n^1), (\dot{x}_n^2, \dot{y}_n^2), \ldots, (\dot{x}_n^{T}, \dot{y}_n^{T})]$, and $\gamma$ and $\beta$ are scaling parameters so the overall variance roughly matches the noise distribution.  Here $\dot{x}_n^{t} = (x_n^{t+1} - x_n^t)$, and $\dot{y}_n^{t} = (y_n^{t+1} - y_n^t)$.  We interpolate occluded values $\dot{x}_n^{t} = (x_n^i - x_n^j)/(i-j)$ where $i$ and $j$ are the next and previous visible points (for occluded points at the end of the sequence, which don't have any such $j$, we simply use 0).  We don't do any special preprocessing for the occlusion indicator; even though it's discrete, we find that the model can still denoise to the discrete values provided that they are scaled appropriately.

\subsection{Diffusion Process}
Following DDPM \cite{ho2020denoising}, we define a forward diffusion process that gradually corrupts the diffusion targets $\mathbf{Z}_0^{\text{diff}}$ with Gaussian noise. The forward process over $\tau = 1, 2, \ldots, S$ diffusion steps is:
\begin{equation}
q(\mathbf{Z}_\tau^{\text{diff}} | \mathbf{Z}_0^{\text{diff}}) = \mathcal{N}(\mathbf{Z}_\tau^{\text{diff}}; \sqrt{\bar{\alpha}_\tau} \mathbf{Z}_0^{\text{diff}}, (1 - \bar{\alpha}_\tau) \mathbf{I}),
\end{equation}
where $\bar{\alpha}_\tau = \prod_{s=1}^\tau \alpha_s$ with $\alpha_s = 1 - \beta_s$ and $\{\beta_s\}_{s=1}^S$ is a linear noise schedule from $\beta_1 = 0.0001$ to $\beta_S = 0.02$.

Our diffusion model, $f_\theta$, learns to reverse this process by predicting the clean diffusable data $\mathbf{Z}_0^{\text{diff}}$ directly. The training objective minimizes the L1 loss:
\begin{equation}
\mathcal{L} = \mathbb{E}_{\mathbf{Z}_0^{\text{diff}}, \tau, \boldsymbol{\epsilon}} \left[ \| \mathbf{Z}_0^{\text{diff}} - f_\theta(\mathbf{Z}_\tau^{\text{diff}}, \mathbf{Z}^{\text{cond}}, \tau) \|_1 \right],
\end{equation}
where $\mathbf{Z}_\tau^{\text{diff}} = \sqrt{\bar{\alpha}_\tau} \mathbf{Z}_0^{\text{diff}} + \sqrt{1 - \bar{\alpha}_\tau} \boldsymbol{\epsilon}$ with $\boldsymbol{\epsilon} \sim \mathcal{N}(\mathbf{0}, \mathbf{1})$, and $\mathbf{Z}^{\text{cond}}=\{\mathbf{I}, \mathbf{X}_{1}, \mathbf{V}_{1:T_c}, \mathbf{O}_{1:T_c}, \mathbf{d}\}$ is conditioning information including the image $\mathbf{I}$ \carl{I reverted this change because DINO is IMO not a property of the diffusion process it's a property of the architecture}, as well as motion and occlusion history and desired displacement, if available.

\medskip
\noindent
\textbf{Diffusion Transformer Architecture.} We now turn to the description of $f_{\theta}$, which predicts the clean tracks given noisy tracks and conditioning information.  We do not assume that tracks are given in any meaningful order or on any grid.  However, similar to~\cite{bharadhwaj2024track2act}, we note that a transformer model, where each token corresponds to a track, can handle the permutation invariance, as long as we include relevant conditioning information within each token that encodes what the track corresponds to.  This design means that the model can easily reason about the full motion forecast for a single point (since everything about a point is encoded within the same point), and yet it can also easily compare and contrast nearby points via attention.  It also means that we can make our network is invariant to the input ordering of the tracks.

Figure~\ref{fig:architecture} shows our overall architecture. Each input token corresponds to a full point trajectory; that is, we construct a token for each track before stacking them into a matrix to pass to the transformer.  Each token contains all per-track conditioning information: image features, and clean history of conditioning velocities and occlusions $\{\textbf{I},\mathbf{X}_1,\mathbf{V}_{1:T_c},\mathbf{O}_{1:T_c}\}$, as well as the noisy diffusion target for the track.  We can then predict the clean data for each track via simple linear projection from the transformer's output. 

We construct a token for the $n$'th point track in the following way. We start with a visual feature derived from $I$, the image frame at time $t=1$.  We extract the full bounding box around the animal plus a 50\% margin, and compute image features from a frozen DINOv3~\cite{simeoni2025dinov3}, which should capture priors about animal parts.  We then extract a feature for the track's initial location $(x^1_n,y^1_n)$ using bilinear interpolation. %
Next, we encode the velocity and occlusion history $(\dot{x}_n^{1:T_{c-1}},\dot{y}_n^{1:T_{c-1}}, \mathbf{O}_n^{1:T_c})$; we embed the velocities $\dot{x}_n^{1:T_{c-1}},\dot{y}_n^{1:T_{c-1}}$ using a sinusoidal embedding and scale by $\gamma$; we keep the occlusions $\mathbf{O}_n^{1:T_c})$ as scalar and multiply by $\beta$.   This component of the token is set to zero in the case where the conditioning is not provided. 
Finally, we add the noisy velocities and occlusion values $\mathbf{Z}^{\text{diff}}_{\tau}=\{\hat{\mathbf{V}},\hat{\mathbf{O}}\}$.   The full token construction is the concatenation of the clean conditioning DINOv3 features, the clean conditioning velocity history embedding and the occlusion history, the noisy velocities, and the noisy occlusions, along the channel dimension: $\mathbf{Z}_n = [\mathbf{Z}^{\text{diff}}_n,\mathbf{f}_n^{\text{DINO}},\mathbf{V}_{1:T_c},\mathbf{O}_{1:T_c}]$.

We project each token to the transformed dimension $D_T$ and add a position encoding.  Unlike sequence models, where the added position encoding is derived from the sequence index, we derive our position encoding from the initial location. $(x_n^1,y_n^1)$.  We use a simple sinusoidal position encoding with length $D_T$ and add it to the track token embedding.   Finally we apply a standard DiT transformer~\cite{peebles2023scalable}, before linearly projecting the final layer to the dimension of each track in $Z^{\text{diff}}$.  

The final conditioning information is global, rather than per-track: the diffusion timestep $\tau$ and optionally the desired total displacement $d$.  We embed these values via a linear embedding, zeroing out the embedding for $d$ in the cases where it is not given, and use adaptive layer norm~\cite{peebles2023scalable} as input directly at each layer of the diffusion model, as is typical for encoding the diffusion timestep in a diffusion transformer.   See appendix~\ref{sec:method_impl_details} for further details on the network architecture, embeddings, and other hyperparameters.

\subsection{Sampling with DDIM}

For efficient inference, we use the DDIM sampling algorithm \cite{songdenoising}, which enables deterministic sampling with fewer steps than the training diffusion process. DDIM defines a non-Markovian forward process that preserves the same marginals $q(\mathbf{Z}_\tau | \mathbf{Z}_0)$ but allows skipping diffusion timesteps during sampling.

Given the model's prediction $\hat{\mathbf{Z}}_0 = f_\theta(\mathbf{Z}_\tau, \tau, \mathbf{d})$ at diffusion timestep $\tau$, we compute the next state $\mathbf{Z}_{\tau-\Delta}$ as:
\begin{align}
\boldsymbol{\epsilon}_\theta &= \frac{\mathbf{Z}_\tau - \sqrt{\bar{\alpha}_\tau} \hat{\mathbf{Z}}_0}{\sqrt{1 - \bar{\alpha}_\tau}}, \\
\mathbf{Z}_{\tau-\Delta} &= \sqrt{\bar{\alpha}_{\tau-\Delta}} \hat{\mathbf{Z}}_0 + \sqrt{1 - \bar{\alpha}_{\tau-\Delta} - \sigma_\tau^2} \boldsymbol{\epsilon}_\theta + \sigma_\tau \boldsymbol{\epsilon},
\end{align}
where $\boldsymbol{\epsilon} \sim \mathcal{N}(0, \mathbf{I})$ and $\sigma_\tau = \eta \sqrt{(1 - \bar{\alpha}_{\tau-\Delta})/(1 - \bar{\alpha}_\tau)} \sqrt{1 - \bar{\alpha}_\tau/\bar{\alpha}_{\tau-\Delta}}$ controls stochasticity. We use deterministic sampling ($\eta = 0$) with 100 diffusion steps instead of the full 1000 training steps, yielding 10$\times$ speedup with minimal quality degradation.
After sampling in velocity space, we convert back to absolute coordinates via cumulative summation: $x_n^t = x_n^1 + \sum_{s=1}^{t-1} v_n^{x,s}$ and $y_n^t = y_n^1 + \sum_{s=1}^{t-1} v_n^{y,s}$ for each point $n$ and trajectory time $t$.

\section{Data Processing for In-the-Wild Animal Videos}

\begin{figure*}[t]
  \centering
    \includegraphics[width=0.99\linewidth]{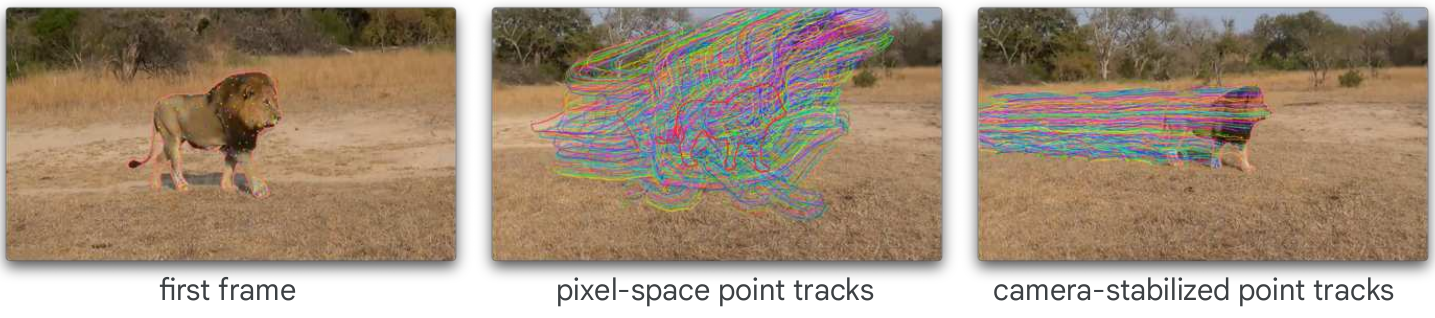}
  \caption{\textbf{Our processed data before and after camera stabilization.} %
  Given a first frame (left), the middle image shows the point tracks in pixel space, where the motion of the animals and the camera (panning, zooming out) are entangled. On the right are our point tracks in camera-stabilized space. We release all of our annotations, including the camera-stabilized point tracks.}
  \label{fig:camera_stab}
  \vspace{-0.9em}
\end{figure*}

Isolating high-fidelity animal motion trajectories from in-the-wild video requires extensive data processing. To this end, we propose a comprehensive pipeline that leverages state-of-the-art perception models to filter data, detect animals, track points on them, and disentangle camera motion from animal motion, ultimately yielding a dataset of stabilized point trajectories suitable for training our animal motion generation model. 

Predicting motion is simpler in world coordinates, as much of the scene remains static and accelerations are predictable from forces~\cite{newton1687principia}.  While there has been much recent work on extracting full 3D coordinates of objects and cameras~\cite{zhang2025d4rt, wang2025vggt}, these algorithms remain unreliable for in-the-wild video of uncommon categories.  However, we note that for naturalist videos, most of the time the cameras are either stationary or shot from a long distance so parallax is minimal.  Therefore, we approximate a world coordinates via simple camera stabilization and normalizing the coordinates with respect to the animal's initial position, as shown in Figure~\ref{fig:camera_stab}.  

\begin{wrapfigure}{r}{.4\textwidth}
  \centering
  \vspace{-2em}
\includegraphics[width=0.98\linewidth]{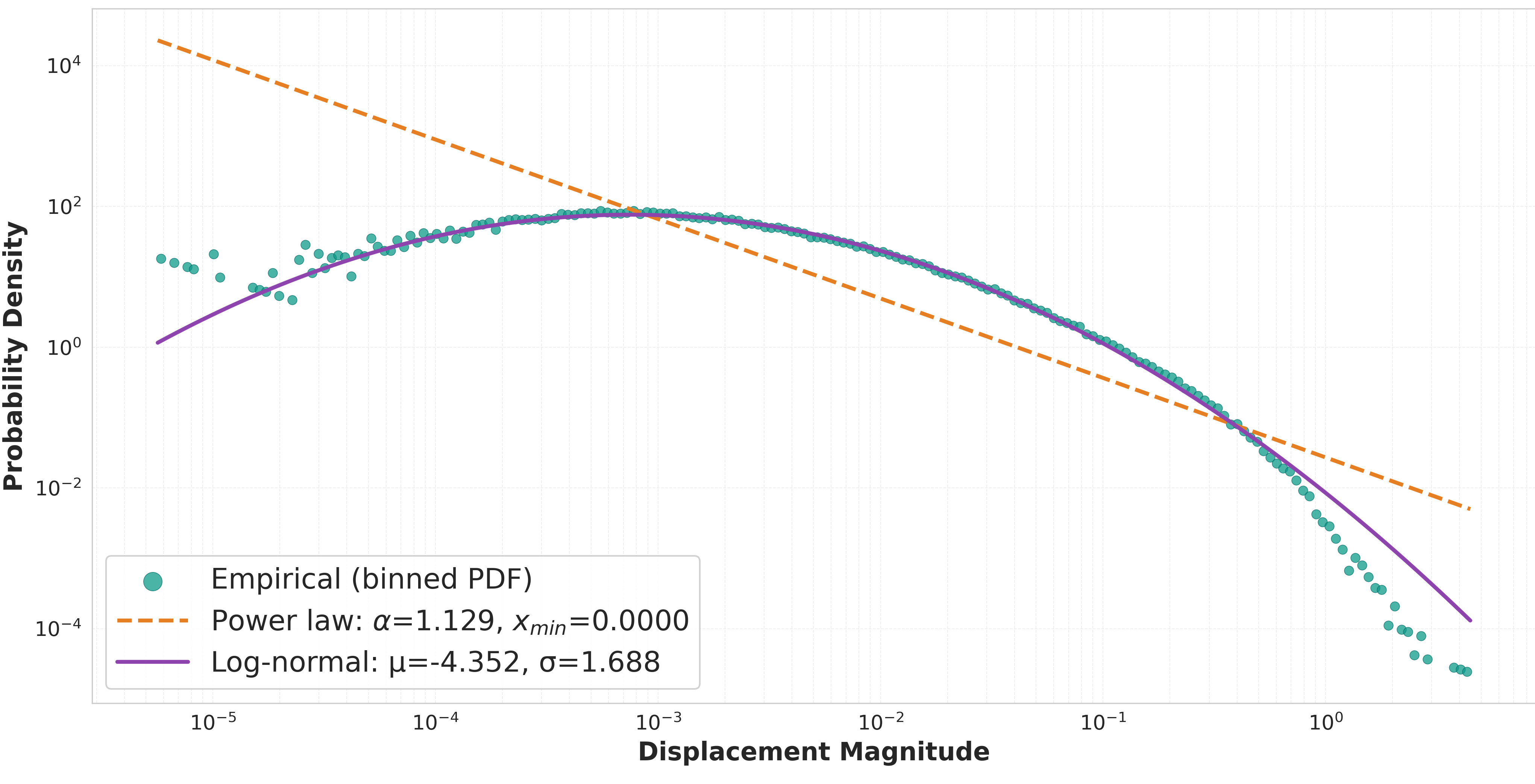}
  \caption{\textbf{Animal motion follows a log normal distribution:} We plot a histogram of animal displacement.  Horizontal axis is a binned log displacement, while vertical axis is log frequency.  We find that log-normal (purple) fits much better than both a power law (orange).}
  \label{fig:log_normal}
  \vspace{-2em}
\end{wrapfigure}

We start by densely tracking points on our videos with BootsTAPIR~\cite{doersch2024bootstap}, and detect individual shots by thresholding the number of points that move smoothly from one frame to the next.  For each shot, we run Grounding-DINO~\cite{liu2023grounding} to get initial animal segments and track them with VideoSAM~\cite{ravi2024sam}.  To perform camera stabilization, we remove tracks that fit mostly inside animal segments, and then perform RANSAC on the rest to estimate a homography $H_t$ for every frame $t$.  

We then extract point tracks within the animal segment, with a bias towards thin structures to ensure they are represented in the resulting tracks.  We then construct training examples by selecting an animal and a particular starting frame $t$.  We extract the input image by taking a bounding box around the segment and expanding it by 50\% on each side.  We transform all other points with respect to this bounding box using the homographies (i.e. multiply each point on frame $t^{\prime}$ by $H_{t}H^{-1}_{t^{\prime}}$).  We then normalize all coordinates with respect to the first bounding box, so that $(0,0)$ corresponds to the upper-left corner and $(1,1)$ the bottom right.  See Supplement~\ref{sec:supp_data_processing} for more details. We release the resulting MammalMotion dataset.

\medskip
\noindent
\textbf{Motion distribution.} 
To validate our motion processing, we compute a histogram of the average displacement (i.e. average distance between start and end for points that remain visible) in Figure~\ref{fig:log_normal}. While one might expect animal motion to follow a power law, we instead find that a log-normal distribution fits far better (i.e., the log displacements are normally distributed).  Such distributions have been found in other datasets of animal motions, e.g. L\'evy flights~\cite{gunner2024high, humphries2010environmental, breed2015apparent}, foraging decisions in rats~\cite{jung2014bursts}, and general spontaneous behavior in animals~\cite{proekt2012scale}, and are suggested to imply that motion magnitude the result of a \textit{multiplicative} interaction of independent factors.  We believe this is the first time such a result has arisen from such a diverse dataset of many different species, and without any painstaking manual annotation or use of tracking devices.

\section{Experimental Setup}

\subsection{Experimental Dataset}

Before processing the data to create MammalMotion, we filter the full 539-hour MammalNet dataset~\cite{chen2023mammalnet}, cutting it down to just under 300 hours. Videos were excluded if they failed to meet minimum requirements for temporal and spatial resolution or displayed a low dynamic range, see~\ref{sec:supp_data_processing} for details.

We evaluate our approach on our filtered \emph{all-species} dataset spanning the entire MammalNet taxonomy, as well as a \emph{Panthera}-only subset comprising lions, tigers, and leopards. For each configuration, we construct evaluation sets by randomly sampling from the validation split with different levels of motion. In the all-species setting, random samples are also drawn using stratified sampling across species $\times$ behavior classes to ensure balanced representation of rare categories.
In contrast, the Panthera-only setting uses uniform random sampling due to its more homogeneous taxonomy. In both cases, we draw even amounts of samples where the animal averages the following amounts of frame-to-frame absolute motion: less than half a pixel, half to $1.5$ pixels, and greater than $1.5$ pixels.
\neerja{refine this}

\subsection{Metrics}

We evaluate our model's performance using a suite of metrics that assess both example-level trajectory accuracy and distribution-level motion. All metrics are computed on predicted trajectories compared against our ground truth.

\paragraph{Distribution-Level Motion Statistics:} we apply several metrics to the overall distributions of predicted trajectories.

\smallskip
\noindent{\textbf{Fréchet Distance (FD).}}
To assess whether our model captures the statistical properties of animal motion, we compute the Fréchet distance~\cite{dowson1982frechet} between predicted and ground truth trajectory distributions. It fits multivariate Gaussian distributions to a set of vectors and compares them.
We compute FD on two representations: first-order differences (velocities), and second-order differences (accelerations), capturing motion dynamics, and motion smoothness, respectively. Following prior work~\cite{walker2025}, we restrict this analysis to individual tracks visible in all predicted frames to ensure complete motion sequences.

\smallskip
\noindent{\textbf{Trajectory Variance.}}
We measure the temporal variance of predicted trajectories $\text{Var}_{\text{pred}} = \text{Var}(\text{flat}(\mathbf{P}^{\text{pred}}))$ where $\mathbf{P}^{\text{pred}} \in \mathbb{R}^{N_{\text{samples}} \times T \times 2}$ is the matrix of all predicted track samples. This captures the diversity and magnitude of motion in generated trajectories. We report this alongside ground truth variance $\text{Var}_{\text{gt}}$ to assess whether the model reproduces natural motion magnitudes.

\smallskip
\noindent{\textbf{Fréchet Video Motion Distance (FVMD).}}
To evaluate temporal coherence, we use the Fréchet Video Motion Distance (FVMD)~\cite{liu2024fr}. FVMD quantifies the discrepancy between the distributions of motion feature vectors, where the features are local histograms of motion orientation and magnitude. 

\paragraph{Example-Level Metrics: } Since diffusion models are stochastic, we follow common practice and report best-of-$K$ metrics by sampling $K=5$ predictions with different random seeds for each test example. For metrics where lower is better (ADE, FDE, VMD), we compute $\min_k \text{metric}_k$ for each example, and average across examples. When higher is better (PWT) we use max. 

\smallskip
\noindent{\textbf{Displacement Error (ADE and FDE).}}
Following standard protocols, we evaluate trajectory accuracy using Average Displacement Error (ADE) and Final Displacement Error (FDE). ADE is the mean squared Euclidean distance between predicted and ground truth trajectories for all visible
points across the predicted timesteps. FDE measures endpoint accuracy at the terminal timestep $T$.

\smallskip
\noindent{\textbf{Points Within Threshold (PWT).}}
As established in point tracking literature~\cite{doersch2023tapir}, we report the fraction of predicted points within pixel-wise distance thresholds of the ground truth $\delta \in \{1, 2, 4, 8, 16\}$, in pixel space, where the input bounding boxes are all resized to (256,256).

\smallskip
\noindent{\textbf{Video Motion Distance (VMD).}} This is a straightforward extension of FVMD to an example-level metric: we compute the same feature vector used for FVMD for both the sample and ground-truth, and report the average Euclidean distance.

\subsection{Baselines}
We first compare our approach against three non-learned baselines. We then also compare our approach with learned baselines ATM and Track2Act. All baselines and our model use $N_{\text{cond}} = 4$ and predict $28$ timesteps at $15$ FPS.

\smallskip
\noindent{\textbf{No-Motion Baseline.}}
The simplest prediction strategy, repeating the last conditioning position for all future timesteps: $\hat{\mathbf{p}}_t = \mathbf{p}_{N_{\text{cond}}-1}$ for $t \geq N_{\text{cond}}$. 

\smallskip
\noindent{\textbf{Constant Velocity Baseline.}}
We estimate a per-point-track velocity from conditioning frames as $\mathbf{v} = (\mathbf{p}_{N_{\text{cond}}-1} - \mathbf{p}_0)/(N_{\text{cond}}-1)$ and linearly extrapolate future positions: $\hat{\mathbf{p}}_t = \mathbf{p}_0 + t \cdot \mathbf{v}$. This provides a simple physics-based predictor assuming constant motion dynamics.

\smallskip
\noindent{\textbf{Oracle Velocity Baseline.}}
Uses ground truth average velocity computed from all of the points on the animal, giving a fair lower bound for the setting of our model that takes ground-truth displacement.

\smallskip
\noindent{\textbf{What Happens Next (WHN).}}WHN~\cite{boduljak2025happens} aims for general-purpose point track forecasting, but the model architecture has a grid constraint that makes it difficult to train on our non-constrained data.  Therefore we apply it zero-shot.

\smallskip
\noindent{\textbf{Any Trajectory Modeling (ATM).}} 
ATM's~\cite{wen2023anypoint} Track Transformer is a regression-based method. Similarly to our method, it treats each point track over time as a token. It masks out future timesteps and learns to regress these coordinates. ATM does not handle visiblity, regresses on absolute xy-coordinates, and can only predict one plausible future. We train this baseline using our Panthera subset, using $N_{\text{cond}} = 4$.

\smallskip
\noindent{\textbf{Track2Act}~\cite{bharadhwaj2024track2act}.}
Most similar to our model, using a diffusion backbone, point track as tokens, and point conditioning setup. We use public Track2Act code and diffuse directly on absolute XY-coordinates, without any positional encoding following the original implementation. We use the\cite{bharadhwaj2024track2act}'s learned ResNet visual features integrated through AdaLN, and use a standard L2 loss. We omit the goal image (unavailable in our setting), condition the model solely on the initial image, and train the model using $N_{\text{cond}} = 4$.

\section{Results}
\begin{figure*}[t]
  \centering
    \includegraphics[width=0.9\linewidth]{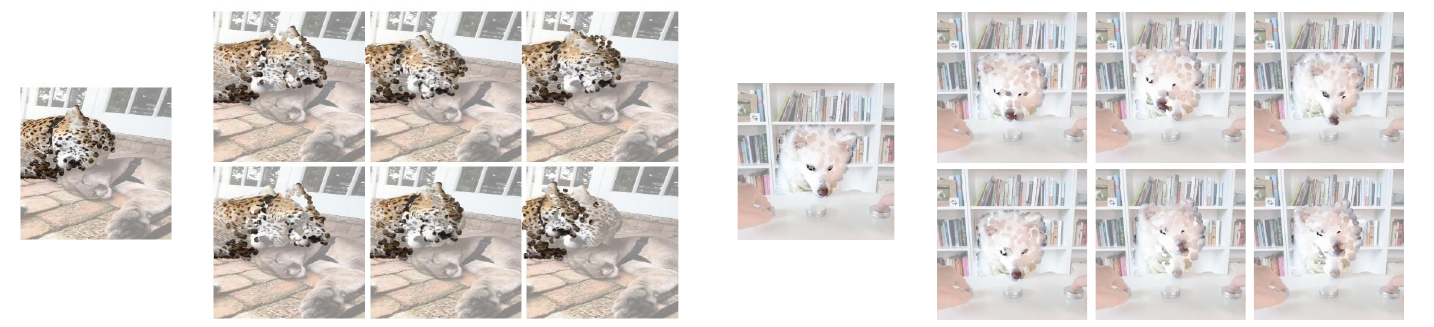}
  \caption{\textbf{Samples from our model} Sampling from our model with different random seeds (each row) and no displacement conditioning. The frame on the left is the input state after the motion history. }
  \label{fig:results}
  \vspace{-0.9em}
\end{figure*}

\begin{figure*}[t]
  \centering
    \includegraphics[width=0.9\linewidth]{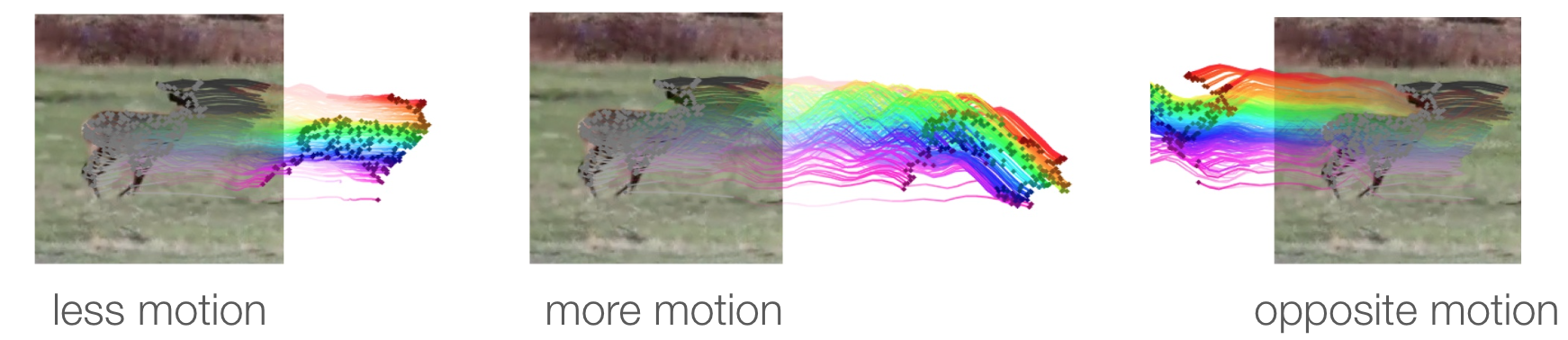}
  \caption{\textbf{Prompting our model with different levels of motion.} We ask the model for less motion (left), more motion (middle), or motion in an opposite direction (right). Grey represents motion history, colors are our predictions.}
  \label{fig:results_motion_prompting}
  \vspace{-0.9em}
\end{figure*}

\begin{figure*}[t]
  \centering
    \includegraphics[width=1\linewidth]{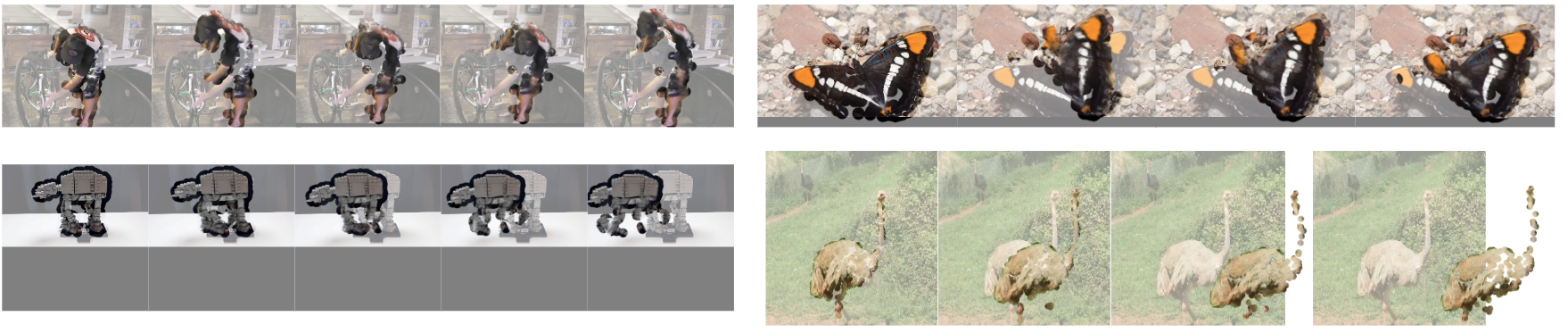}
  \caption{\textbf{Out-of-distribution:} Our model \textit{generalizes} to humans and non-mammals.}
  \label{fig:results_OOD}
  \vspace{-0.9em}
\end{figure*}

\noindent
\textbf{Samples from our model.}
Figure~\ref{fig:teaser} shows qualitative results which exhibit convincing forecasts across a variety of behaviors: e.g. the lion's legs follows natural articulation, the jaguar scouts out prey, the alpaca grazes naturally. This also works for rare animal categories.  Figure~\ref{fig:results} shows the diversity that our model produces with only different seeds: we see different frequencies of the grooming behavior for the jaguar and the dog's head moves different directions.  Fig.~\ref{fig:results_motion_prompting} demonstrates prompting our model for more motion, less motion, or motion in a different direction, and the model produces plausible behaviors consistent with these motions.  Static visualizations do not do justice to motion accuracy; we urge our readers to check the results \href{https://motion-forecasting.github.io/}{webpage} for better visualizations of these effects.

\medskip
\noindent
\textbf{Out-of-distribution examples.}
Fig.~\ref{fig:results_OOD} displays qualitative results of our model on out-of-distribution data. We note that the MammalNet dataset does have videos that contain humans and other types of animals; the ostrich on the bottom right was found in our validation set, so this type of generalization is not surprising. However, the Lego robot (bottom left) and butterfly (top right) are unlike the expected MammalNet data distribution, but still observe physically plausible motion. 

\begin{figure*}[t]
  \centering
    \includegraphics[trim=1cm 6cm 1cm 6cm, clip, width=\textwidth]{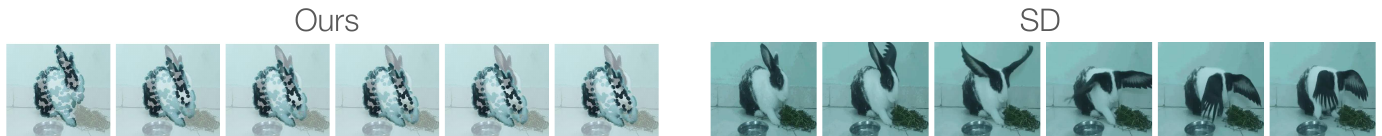}
  \caption{\textbf{Our Model vs. Stable Diffusion.} Our approach can model the behavior of less common animals in our dataset such as hares, while conventional video models struggle with these animals.}
  \label{fig:ours_vs_video}
  \vspace{-0.9em}
\end{figure*}

\medskip
\noindent
\textbf{Comparison with Video Generation Models.}
Figure~\ref{fig:ours_vs_video} displays a side-by-side comparison with Stable Video Diffusion~\cite{blattmann2023stable}. As video models are forced to simulate information beyond motion---textures, lighting, and more---they can struggle to produce physically realistic motions. In contrast, because we model trajectory tokens directly, we can produce more realistic biological behavior with less compute and data. These extend the findings of \cite{boduljak2025happens} on synthetic data of rigid objects to the in-the-wild nonrigid setting. E.g., our model is able to forecast the foraging behavior of a hare even though hares only constitute $0.39\%$ of the training data. The video model not only struggles to model this behavior but even fails to model the animal itself, morphing the ears of the hare into wings.

\subsection{Quantitative Comparison with Baselines}
Results comparing our method with baselines are presented in table~\ref{tab:all_data_results_distribution} and table~\ref{tab:results_all_data_per_example} for our full MammalMotion dataset and table~\ref{tab:panthera_results_distribution} and table~\ref{tab:results_panthera_per_example} for the Panthera genus subset. Each table shows models trained (for learned baselines) and evaluated either on the full dataset or restricted to the \textit{Panthera} subset.

Simple baselines, such as the no-motion and constant-velocity models, can achieve competitive performance on the combined dataset due to the prevalence of low-motion examples. Tables~\ref{tab:results_all_data_dist} and~\ref{tab:results_all_data} in the appendix show results for all motion buckets.  We see that on the low and medium motion data subsets (where an animal might be eating, sleeping, or just not moving), an unlearned baseline of just predicting no motion outperforms the other baselines on several metrics.
These simple baselines, however, fail for higher motion, and particularly for FVMD which accurately scores motion statistics.  Interestingly, WHN gives an accurate acceleration distribution despite not being trained on this data, yet fails to estimate overall velocity and other statistics well (qualitatively it gives low-motion, jittery predictions that don't match animal skeletons). 

More advanced baselines, including ATM and Track2Act--both retrained on the \textit{Panthera} subset, with Track2Act additionally trained on the full dataset--give predictions that are closer to the ground truth in terms of endpoint error and velocity statistics. These methods actually perform worse in terms of acceleration and point-level accuracy, suggesting they learn overall motion but miss motion details.

In contrast, our method, trained from scratch on MammalMotion data (or the Panthera subset), substantially outperforms others in prediction accuracy on every metric.  Furthermore, our method can take the true velocity as conditioning to improve results even further, even though for many metrics simply using the oracle velocity provides little boost.  

Comparing our model trained on the Panthera data subset vs the full dataset, our model trained on the full data is substantially better, e.g. FVMD for high motion examples falls from 84.8 to 49.3 and PWT rises from 20.6 to 26.0.  This isn't because the full dataset is easier than the Panthera subset; other baselines actually perform similarly or worse on these metrics.  Instead, this suggests that training on the full dataset improves performance due to transfer between species.

\begin{table}[t]
\begin{minipage}{0.55\textwidth}
\centering
\scriptsize %
\caption{Distribution-level quantitative results on \textbf{All Data}. FD values are multiplied by $10^3$; Variance values are multiplied by $10^5$; FVMD values are divided by $10^3$. V and A refer to velocity and acceleration respectively. Best results in \textbf{bold}, second best \underline{underlined}. $\uparrow$ indicates higher is better; $\downarrow$ indicates lower is better.}
\label{tab:all_data_results_distribution}
\resizebox{\linewidth}{!}{
\begin{tabular}{@{}ll ccccc@{}}
\toprule
& Method & FD(V)$\downarrow$ & FD(A)$\downarrow$ & Var(V) & Var(A) & FVMD$\downarrow$ \\
\midrule
\multirow{9}{*}{\rotatebox{90}{High motion}}
    & GT & - & - & 31.5 & 8.94 & - \\
    & No motion & 27.1 & 7.51 & 0 & 0 & 481.99 \\
    & Constant vel & 13.7 & 7.51 & 23.8 & 0 & 210.47 \\
    & WHN & 25.2 & \textbf{3.19} & 1.1 & 3.34 & 280.77 \\
    & Track2Act & \underline{11.8} & 4.81 & 9.89 & 1.63 & \underline{76.28} \\
    & Ours (uncond) & \textbf{8.96} & \underline{3.74} & 13.1 & 1.68 & \textbf{49.30} \\
    \cmidrule(lr){2-7}
    & Oracle vel & 12.1 & 7.51 & 19.8 & 0 & 326.80 \\
    & Ours (cond) & \textbf{4.86} & \textbf{3.33} & 28.3 & 2.14 & \textbf{40.24} \\
\cmidrule(lr){1-7}
\multirow{9}{*}{\rotatebox{90}{Combined}}
   & GT & - & - & 5.41 & 1.82 & - \\
    & No motion & 4.66 & 1.53 & 0 & 0 & 204.14 \\
    & Constant vel & \underline{2.59} & 1.53 & 4.11 & 0 & 89.77 \\
    & WHN & 5.34 & \textbf{0.691} & 1.23 & 3.7 & 94.7 \\
    & Track2Act & \underline{2.52} & 1.09 & 2.37 & 0.516 & \underline{26.17} \\
    & Ours (uncond) & \textbf{1.96} & \underline{0.877} & 2.94 & 0.453 & \textbf{17.0} \\
    \cmidrule(lr){2-7}
    & Oracle vel & 2.26 & 1.53 & 3.15 & 0 & 185.62 \\
    & Ours (cond) & \textbf{1.07} & \textbf{0.778} & 6.26 & 0.57 & \textbf{14.38} \\
\bottomrule

\end{tabular}
}

\end{minipage}\hfill %
\begin{minipage}{0.42\textwidth}
\centering
\scriptsize %
\caption{Example-level quantitative results on \textbf{All Data}. Best results in \textbf{bold}, second best \underline{underlined}. For non-learned baselines we report single-sample metrics; for WHN, Track2Act, and Ours we report best of $K=5$.}
\label{tab:results_all_data_per_example}
\resizebox{\textwidth}{!}{%
\begin{tabular}{l l cccc}
\toprule
                                             & Method        & ADE$\downarrow$  & FDE$\downarrow$  & VMD$\downarrow$  & PWT$\uparrow$ \\
\midrule
\multirow{8}{*}{\rotatebox{90}{High motion}}  & No motion & 0.325 & 0.596 & 6.50 & 12.44\% \\
    & Constant vel & 0.286 & 0.591 & 5.02 & 11.94\% \\
    & WHN & 0.262 & 0.538 & 5.74 & 11.62\% \\
    & Track2Act & \underline{0.136} & \underline{0.294} & \underline{4.50} & \underline{21.79\%} \\
    & Ours (uncond) & \textbf{0.119} & \textbf{0.275} & \textbf{4.33} & \textbf{26.01\%} \\
    \cmidrule(lr){2-6}
    & Oracle vel & 0.110 & 0.156 & 7.04 & 14.70\% \\
    & Ours (cond) & \textbf{0.068} & \textbf{0.103} & \textbf{4.25} & \textbf{31.50\%} \\
\cmidrule(lr){1-6}
\multirow{8}{*}{\rotatebox{90}{Combined}}     & No motion & 0.099 & 0.180 & 4.82 & 53.94\% \\
    & Constant vel & 0.104 & 0.215 & 5.02 & 41.15\% \\
    & WHN & 0.105 & 0.200 & 4.85 & 29.92\% \\
     & Track2Act & \underline{0.053} & \underline{0.110} & \underline{3.48} & \underline{56.59\%} \\
    & Ours (uncond) & \textbf{0.046} & \textbf{0.102} & \textbf{3.31} & \textbf{60.01\%} \\
    \cmidrule(lr){2-6}
    & Oracle vel & 0.042 & 0.058 & 5.57 & 51.74\% \\
    & Ours (cond) & \textbf{0.028} & \textbf{0.042} & \textbf{3.26} & \textbf{63.48\%} \\
\bottomrule
\end{tabular}%
}
\end{minipage}
\end{table}

\begin{table}[t]
\begin{minipage}{0.55\textwidth}
\centering
\scriptsize %
\caption{Distribution-level quantitative results on \textbf{Panthera}. FD values are multiplied by $10^3$; Variance values are multiplied by $10^5$; FVMD values are divided by $10^3$. V and A refer to velocity and acceleration respectively. Best results in \textbf{bold}, second best \underline{underlined}. $\uparrow$ indicates higher is better; $\downarrow$ indicates lower is better.}
\label{tab:panthera_results_distribution}
\resizebox{\linewidth}{!}{
\begin{tabular}{@{}ll ccccc@{}}
\toprule
& Method & FD(V)$\downarrow$ & FD(A)$\downarrow$ & Var(V) & Var(A) & FVMD$\downarrow$ \\
\midrule
\multirow{9}{*}{\rotatebox{90}{High motion}}
    & GT & - & - & 29.5 & 10.8 & - \\
    & No motion & 16.6 & 5.61 & 0 & 0 & 335.41 \\
    & Constant vel & 7.49 & 5.61 & 37.3 & 0 & 149.52 \\
    & WHN & 15.2 & \textbf{3.27} & 1.37 & 4.11 & 247.56 \\
    & ATM & 6.52 & 6.18 & 10 & 6.99 & 112.71 \\
    & Track2Act & \underline{6.32} & 5.06 & 8.01 & 0.45 & \underline{104.85} \\
    & Ours (uncond) & \textbf{3.71} & \underline{4.3} & 12.8 & 1.02 & \textbf{84.79} \\
    \cmidrule(lr){2-7}
    & Oracle vel & 5.7 & 5.61 & 20.9 & 0 & 218.73 \\
    & Ours (cond) & \textbf{2.82} & \textbf{4.19} & 16.6 & 1.18 & \textbf{79.38} \\
\cmidrule(lr){1-7}
\multirow{9}{*}{\rotatebox{90}{Combined}}
    & GT & - & - & 6.93 & 2.66 & - \\
    & No motion & 3.77 & 1.38 & 0 & 0 & 149.53 \\
    & Constant vel & 1.86 & 1.38 & 8.58 & 0 & 62.51 \\
    & WHN & 3.37 & \underline{1.12} & 1.43 & 4.29 & 86.89 \\
    & ATM & 1.49 & 1.4 & 2.42 & 1.75 & \underline{35.50} \\
    & Track2Act & \underline{1.43} & 1.21 & 1.89 & .121 & 38.44 \\
    & Ours (uncond) & \textbf{.874} & \textbf{1.05} & 2.94 & .226 & \textbf{24.82} \\
    \cmidrule(lr){2-7}
    & Oracle vel & 1.43 & 1.38 & 4.73 & 0 & 118.61 \\
    & Ours (cond) & \textbf{.679} & \textbf{1.02} & 3.73 & .262 & \textbf{24.90} \\
\bottomrule

\end{tabular}
}

\end{minipage}\hfill %
\begin{minipage}{0.42\textwidth}

\centering
\scriptsize %
\caption{Example-level quantitative results on \textbf{Panthera}. Best results in \textbf{bold}, second best \underline{underlined}. For non-learned baselines and ATM (single output), we report single-sample metrics; for WHN, Track2Act, and Ours we report best of $K=5$.}
\label{tab:results_panthera_per_example}
\resizebox{\textwidth}{!}{%
\begin{tabular}{l l cccc}
\toprule
                                             & Method        & ADE$\downarrow$  & FDE$\downarrow$  & VMD$\downarrow$  & PWT$\uparrow$ \\
\midrule
\multirow{8}{*}{\rotatebox{90}{High motion}} & No motion     &         .211      &         .393      &         5.51      &         13.22\%           \\
                                             & Constant vel  &         .193      &         .413      & \underline{4.91}  & \underline{16.73\%}       \\
                                             & WHN           &         .215      &         .393      &         5.82      &         10.24\%           \\
                                             & ATM           &         .143      &         .262      &         5.95      &         16.31\%           \\
                                             & Track2Act     & \underline{.135}  & \underline{.245}  &         5.04      &         16.72\%           \\
                                             & Ours (uncond) & \textbf{.107}     & \textbf{.209}     & \textbf{4.77}     & \textbf{20.68\%}          \\
\cmidrule(lr){2-6}
                                             & Oracle vel    &         .082      & \textbf{.095}     &         6.03      &         17.05\%           \\
                                             & Ours (cond)   & \textbf{.067}     &         .097      & \textbf{4.61}     & \textbf{27.31\%}          \\
\cmidrule(lr){1-6}
\multirow{8}{*}{\rotatebox{90}{Combined}}    & No motion     &         .076      &         .138      &         4.02      & \underline{54.55\%}       \\
                                             & Constant vel  &         .079      &         .164      &         4.33      &         46.16\%           \\
                                             & WHN           &         .086      &         .146      &         5.05      &         30.43\%           \\
                                             & ATM           &         .057      &         .101      &         4.48      &         48.39\%           \\
                                             & Track2Act     & \underline{.053}  & \underline{.092}  & \underline{3.81}  &         51.13\%           \\
                                             & Ours (uncond) & \textbf{.042}     & \textbf{.078}     & \textbf{3.62}     & \textbf{58.11\%}          \\
\cmidrule(lr){2-6}
                                             & Oracle vel    &         .035      &         .042      &         4.51      &         53.14\%           \\
                                             & Ours (cond)   & \textbf{.028}     & \textbf{.039}     & \textbf{3.55}     & \textbf{61.67\%}          \\
\bottomrule
\end{tabular}%
}
\end{minipage}
\end{table}

\section{Conclusion}

Our paper demonstrates a data-efficient and yet highly-general framework for motion forecasting.  The critical ingredient is a token representation based on individual trajectories, using DINO features and track velocities as input, and a transformer-based diffusion network to model them.  Our results show strong performance for animal forecasting across many species, producing diverse samples and allowing control of output velocity.  Results even transfer out of distribution.  We believe this model can serve as a foundation for more general modeling of arbitrary object motion, and extensions may extract knowledge about animal species that are useful to ecologists, potentially saving the incredible labor that's currently expended for behavior modeling~\cite{anderson2014toward}.  

\section{Acknowledgments}

We thank Noah Snavely for challenging us with general motion forecasting over a lovely Parisian lunch. We thank 
Andrew Zisserman,
Drew Purves,
Aleksander Holynski,
Linyi Jin,
Sander Dieleman,
Mark Hamilton,
and Jathushan Rajasegeran
for helpful discussions and feedback. This work was supported by ONR MURI N00014-21-1-280, and a NSF Graduate Fellowship to NT.

\bibliographystyle{splncs04}
\bibliography{main}

\begin{thebibliography}{10}
\providecommand{\url}[1]{\texttt{#1}}
\providecommand{\urlprefix}{URL }
\providecommand{\doi}[1]{https://doi.org/#1}

\bibitem{anderson2014toward}
Anderson, D.J., Perona, P.: Toward a science of computational ethology. Neuron  \textbf{84}(1),  18--31 (2014)

\bibitem{atkinson2004emotion}
Atkinson, A.P., Dittrich, W.H., Gemmell, A.J., Young, A.W.: Emotion perception from dynamic and static body expressions in point-light and full-light displays. Perception  \textbf{33}(6),  717--746 (2004)

\bibitem{bansalvideophy}
Bansal, H., Lin, Z., Xie, T., Zong, Z., Yarom, M., Bitton, Y., Jiang, C., Sun, Y., Chang, K.W., Grover, A.: Videophy: Evaluating physical commonsense for video generation. In: The Thirteenth International Conference on Learning Representations (2025)

\bibitem{bharadhwaj2025gen2act}
Bharadhwaj, H., Dwibedi, D., Gupta, A., Tulsiani, S., Doersch, C., Xiao, T., Shah, D., Xia, F., Sadigh, D., Kirmani, S.: Gen2act: Human video generation in novel scenarios enables generalizable robot manipulation. In: Conference on Robot Learning. pp. 3936--3951. PMLR (2025)

\bibitem{bharadhwaj2024track2act}
Bharadhwaj, H., Mottaghi, R., Gupta, A., Tulsiani, S.: Track2act: Predicting point tracks from internet videos enables generalizable robot manipulation. In: European Conference on Computer Vision (ECCV) (2024)

\bibitem{blattmann2023stable}
Blattmann, A., Dockhorn, T., Kulal, S., Mendelevitch, D., Kilian, M., Lorenz, D., Levi, Y., English, Z., Voleti, V., Letts, A., et~al.: Stable video diffusion: Scaling latent video diffusion models to large datasets. arXiv preprint arXiv:2311.15127  (2023)

\bibitem{boduljak2025happens}
Boduljak, G., Karazija, L., Laina, I., Rupprecht, C., Vedaldi, A.: What happens next? anticipating future motion by generating point trajectories. arXiv preprint arXiv:2509.21592  (2025)

\bibitem{breed2015apparent}
Breed, G.A., Severns, P.M., Edwards, A.M.: Apparent power-law distributions in animal movements can arise from intraspecific interactions. Journal of the Royal Society Interface  \textbf{12}(103) (2015)

\bibitem{chefer2025}
Chefer, H., Singer, U., Zohar, A., Kirstain, Y., Polyak, A., Taigman, Y., Wolf, L., Sheynin, S.: Videojam: Joint appearance-motion representations for enhanced motion generation in video models. In: Forty-second International Conference on Machine Learning

\bibitem{chen2023mammalnet}
Chen, J., Hu, M., Coker, D.J., Berumen, M.L., Costelloe, B., Beery, S., Rohrbach, A., Elhoseiny, M.: Mammalnet: A large-scale video benchmark for mammal recognition and behavior understanding. In: Proceedings of the IEEE/CVF conference on computer vision and pattern recognition. pp. 13052--13061 (2023)

\bibitem{Chen_2025_ICCV}
Chen, Y., Li, P., Huang, Y., Yang, J., Chen, K., Wang, L.: Ec-flow: Enabling versatile robotic manipulation from action-unlabeled videos via embodiment-centric flow. In: Proceedings of the IEEE/CVF International Conference on Computer Vision (ICCV). pp. 11958--11968 (October 2025)

\bibitem{clark2019adversarial}
Clark, A., Donahue, J., Simonyan, K.: Adversarial video generation on complex datasets. arXiv preprint arXiv:1907.06571  (2019)

\bibitem{cutting1977recognizing}
Cutting, J.E., Kozlowski, L.T.: Recognizing friends by their walk: Gait perception without familiarity cues. Bulletin of the psychonomic society  \textbf{9}(5),  353--356 (1977)

\bibitem{doersch2024bootstap}
Doersch, C., Luc, P., Yang, Y., Gokay, D., Koppula, S., Gupta, A., Heyward, J., Rocco, I., Goroshin, R., Carreira, J., et~al.: Bootstap: Bootstrapped training for tracking-any-point. In: Proceedings of the Asian Conference on Computer Vision. pp. 3257--3274 (2024)

\bibitem{doersch2023tapir}
Doersch, C., Yang, Y., Vecerik, M., Gokay, D., Gupta, A., Aytar, Y., Carreira, J., Zisserman, A.: Tapir: Tracking any point with per-frame initialization and temporal refinement. In: Proceedings of the IEEE/CVF International Conference on Computer Vision. pp. 10061--10072 (2023)

\bibitem{dowson1982frechet}
Dowson, D.C., Landau, B.: The fr{\'e}chet distance between multivariate normal distributions. Journal of multivariate analysis  \textbf{12}(3),  450--455 (1982)

\bibitem{fox1982perception}
Fox, R., McDaniel, C.: The perception of biological motion by human infants. Science  \textbf{218}(4571),  486--487 (1982)

\bibitem{flip}
Gao, C., Zhang, H., Xu, Z., Cai, Z., Shao, L.: Flip: Flow-centric generative planning for general-purpose manipulation tasks. arXiv preprint arXiv:2412.08261  (2024), \url{https://arxiv.org/abs/2412.08261}

\bibitem{google_veo3_2025}
{Google DeepMind}: Veo 3 technical report. Tech. rep., Google DeepMind (2025)

\bibitem{greff2022kubric}
Greff, K., Belletti, F., Beyer, L., Doersch, C., Du, Y., Duckworth, D., Fleet, D.J., Gnanapragasam, D., Golemo, F., Herrmann, C., et~al.: Kubric: A scalable dataset generator. In: Proceedings of the IEEE/CVF conference on computer vision and pattern recognition. pp. 3749--3761 (2022)

\bibitem{grossman2000brain}
Grossman, E., Donnelly, M., Price, R., Pickens, D., Morgan, V., Neighbor, G., Blake, R.: Brain areas involved in perception of biological motion. Journal of cognitive neuroscience  \textbf{12}(5),  711--720 (2000)

\bibitem{gu2023seer}
Gu, X., Wen, C., Ye, W., Song, J., Gao, Y.: Seer: Language instructed video prediction with latent diffusion models. arXiv preprint arXiv:2303.14897  (2023)

\bibitem{gunner2024high}
Gunner, R., Wilson, R., Lurgi, M., Borger, L., Redcliffe, J., Shepard, E., Holton, M., Crofoot, M., Alagaili, A., Andrzejaczek, S., et~al.: High resolution data reveal fundamental steps and turning points in animal movements  (2024)

\bibitem{gupta2018social}
Gupta, A., Johnson, J., Fei-Fei, L., Savarese, S., Alahi, A.: Social gan: Socially acceptable trajectories with generative adversarial networks. In: Proceedings of the IEEE conference on computer vision and pattern recognition. pp. 2255--2264 (2018)

\bibitem{gupta2024walt}
Gupta, A., Yu, L., Sohn, K., Gu, X., Hahn, M., Fei-Fei, L., Essa, I., Jiang, L., Lezama, J.: Photorealistic video generation with diffusion models. In: ECCV (2024)

\bibitem{ho2020denoising}
Ho, J., Jain, A., Abbeel, P.: Denoising diffusion probabilistic models. Advances in neural information processing systems  \textbf{33},  6840--6851 (2020)

\bibitem{hoppe2022diffusion}
H{\"o}ppe, T., Mehrjou, A., Bauer, S., Nielsen, D., Dittadi, A.: Diffusion models for video prediction and infilling. arXiv preprint arXiv:2206.07696  (2022)

\bibitem{humphries2010environmental}
Humphries, N.E., Queiroz, N., Dyer, J.R., Pade, N.G., Musyl, M.K., Schaefer, K.M., Fuller, D.W., Brunnschweiler, J.M., Doyle, T.K., Houghton, J.D., et~al.: Environmental context explains l{\'e}vy and brownian movement patterns of marine predators. Nature  \textbf{465}(7301),  1066--1069 (2010)

\bibitem{johansson1973visual}
Johansson, G.: Visual perception of biological motion and a model for its analysis. Perception \& psychophysics  \textbf{14}(2),  201--211 (1973)

\bibitem{jung2014bursts}
Jung, K., Jang, H., Kralik, J.D., Jeong, J.: Bursts and heavy tails in temporal and sequential dynamics of foraging decisions. PLoS computational biology  \textbf{10}(8),  e1003759 (2014)

\bibitem{kang2025far}
Kang, B., Yue, Y., Lu, R., Lin, Z., Zhao, Y., Wang, K., Huang, G., Feng, J.: How far is video generation from world model: A physical law perspective. In: International Conference on Machine Learning. pp. 28991--29017. PMLR (2025)

\bibitem{karaev2025cotracker3}
Karaev, N., Makarov, Y., Wang, J., Neverova, N., Vedaldi, A., Rupprecht, C.: Cotracker3: Simpler and better point tracking by pseudo-labelling real videos. In: Proceedings of the IEEE/CVF International Conference on Computer Vision. pp. 6013--6022 (2025)

\bibitem{karaev2024cotracker}
Karaev, N., Rocco, I., Graham, B., Neverova, N., Vedaldi, A., Rupprecht, C.: Cotracker: It is better to track together. In: European conference on computer vision. pp. 18--35. Springer (2024)

\bibitem{kitani2012activity}
Kitani, K.M., Ziebart, B.D., Bagnell, J.A., Hebert, M.: Activity forecasting. In: European conference on computer vision. pp. 201--214. Springer (2012)

\bibitem{kozlowski1977recognizing}
Kozlowski, L.T., Cutting, J.E.: Recognizing the sex of a walker from a dynamic point-light display. Perception \& psychophysics  \textbf{21}(6),  575--580 (1977)

\bibitem{lauer2022multi}
Lauer, J., Zhou, M., Ye, S., Menegas, W., Schneider, S., Nath, T., Rahman, M.M., Di~Santo, V., Soberanes, D., Feng, G., et~al.: Multi-animal pose estimation, identification and tracking with deeplabcut. Nature Methods  \textbf{19}(4),  496--504 (2022)

\bibitem{lee2018stochastic}
Lee, A.X., Zhang, R., Ebert, F., Abbeel, P., Finn, C., Levine, S.: Stochastic adversarial video prediction. arXiv preprint arXiv:1804.01523  (2018)

\bibitem{lee2017desire}
Lee, N., Choi, W., Vernaza, P., Choy, C.B., Torr, P.H., Chandraker, M.: Desire: Distant future prediction in dynamic scenes with interacting agents. In: Proceedings of the IEEE conference on computer vision and pattern recognition. pp. 336--345 (2017)

\bibitem{Li_2025_ICCV}
Li, S., Liu, C., Xu, X., Yeo, S.Y., Yang, X.: Future-aware interaction network for motion forecasting. In: Proceedings of the IEEE/CVF International Conference on Computer Vision (ICCV). pp. 7505--7515 (October 2025)

\bibitem{liu2024fr}
Liu, J., Qu, Y., Yan, Q., Zeng, X., Wang, L., Liao, R.: Fr$\backslash$'echet video motion distance: A metric for evaluating motion consistency in videos. arXiv preprint arXiv:2407.16124  (2024)

\bibitem{liu2023grounding}
Liu, S., Zeng, Z., Ren, T., Li, F., Zhang, H., Yang, J., Li, C., Yang, J., Su, H., Zhu, J., et~al.: Grounding dino: Marrying dino with grounded pre-training for open-set object detection. arXiv preprint arXiv:2303.05499  (2023)

\bibitem{Liu_2021_ICCV}
Liu, Z., Su, P., Wu, S., Shen, X., Chen, H., Hao, Y., Wang, M.: Motion prediction using trajectory cues. In: Proceedings of the IEEE/CVF International Conference on Computer Vision (ICCV). pp. 13299--13308 (October 2021)

\bibitem{loper2023smpl}
Loper, M., Mahmood, N., Romero, J., Pons-Moll, G., Black, M.J.: Smpl: A skinned multi-person linear model. In: Seminal Graphics Papers: Pushing the Boundaries, Volume 2, pp. 851--866 (2023)

\bibitem{lorenz1935kumpan}
Lorenz, K.: Der kumpan in der umwelt des vogels. der artgenosse als ausl{\"o}sendes moment sozialer verhaltungsweisen. Journal f{\"u}r Ornithologie. Beiblatt.(Leipzig)  (1935)

\bibitem{lorenz1938taxis}
Lorenz, K., Tinbergen, N.: Taxis und instinkthandlung in der eirollbewegung der graugans. Zeitschrift f{\"u}r Tierpsychologie  (1938)

\bibitem{mangalam2020not}
Mangalam, K., Girase, H., Agarwal, S., Lee, K.H., Adeli, E., Malik, J., Gaidon, A.: It is not the journey but the destination: Endpoint conditioned trajectory prediction. In: European conference on computer vision. pp. 759--776. Springer (2020)

\bibitem{mathis2018deeplabcut}
Mathis, A., Mamidanna, P., Cury, K.M., Abe, T., Murthy, V.N., Mathis, M.W., Bethge, M.: Deeplabcut: markerless pose estimation of user-defined body parts with deep learning. Nature neuroscience  \textbf{21}(9),  1281--1289 (2018)

\bibitem{moon2024visiontrap}
Moon, S., Woo, H., Park, H., Jung, H., Mahjourian, R., Chi, H.g., Lim, H., Kim, S., Kim, J.: Visiontrap: Vision-augmented trajectory prediction guided by textual descriptions. In: European Conference on Computer Vision. pp. 361--379. Springer (2024)

\bibitem{nath2019using}
Nath, T., Mathis, A., Chen, A.C., Patel, A., Bethge, M., Mathis, M.W.: Using deeplabcut for 3d markerless pose estimation across species and behaviors. Nature protocols  \textbf{14}(7),  2152--2176 (2019)

\bibitem{newton1687principia}
Newton, I.: Philosophiae Naturalis Principia Mathematica (1687)

\bibitem{niu2025pre}
Niu, D., Sharma, Y., Xue, H., Biamby, G., Zhang, J., Ji, Z., Darrell, T., Herzig, R.: Pre-training auto-regressive robotic models with 4d representations. arXiv preprint arXiv:2502.13142  (2025)

\bibitem{noronhaquadforecaster}
Noronha, I., Chowdhury, A., Bharti, S., Kaur, U.: Quadforecaster: Diffusion-based quadruped pose prediction for animal communication analysis. In: The Thirty-Ninth Annual Conference on Neural Information Processing Systems workshop: AI for non-human animal communication

\bibitem{openai-2024}
OpenAI: {Sora} (12 2024), \url{https://openai.com/sora/}

\bibitem{9294028}
Oprea, S., Martinez-Gonzalez, P., Garcia-Garcia, A., Castro-Vargas, J.A., Orts-Escolano, S., Garcia-Rodriguez, J., Argyros, A.: A review on deep learning techniques for video prediction. IEEE Transactions on Pattern Analysis and Machine Intelligence  \textbf{44}(6),  2806--2826 (2022). \doi{10.1109/TPAMI.2020.3045007}

\bibitem{peebles2023scalable}
Peebles, W., Xie, S.: Scalable diffusion models with transformers. In: Proceedings of the IEEE/CVF international conference on computer vision. pp. 4195--4205 (2023)

\bibitem{pereira2022sleap}
Pereira, T.D., Tabris, N., Matsliah, A., Turner, D.M., Li, J., Ravindranath, S., Papadoyannis, E.S., Normand, E., Deutsch, D.S., Wang, Z.Y., et~al.: Sleap: A deep learning system for multi-animal pose tracking. Nature methods  \textbf{19}(4),  486--495 (2022)

\bibitem{polajnar2024wing}
Polajnar, J., Kvinikadze, E., Harley, A.W., Malenovsk{\`y}, I.: Wing buzzing as a mechanism for generating vibrational signals in psyllids (hemiptera: Psylloidea). Insect science  \textbf{31}(5),  1466--1476 (2024)

\bibitem{proekt2012scale}
Proekt, A., Banavar, J.R., Maritan, A., Pfaff, D.W.: Scale invariance in the dynamics of spontaneous behavior. Proceedings of the National Academy of Sciences  \textbf{109}(26),  10564--10569 (2012)

\bibitem{ranzato2014video}
Ranzato, M., Szlam, A., Bruna, J., Mathieu, M., Collobert, R., Chopra, S.: Video (language) modeling: a baseline for generative models of natural videos. arXiv preprint arXiv:1412.6604  (2014)

\bibitem{ravi2024sam}
Ravi, N., Gabeur, V., Hu, Y.T., Hu, R., Ryali, C., Ma, T., Khedr, H., R{\"a}dle, R., Rolland, C., Gustafson, L., et~al.: Sam 2: Segment anything in images and videos. arXiv preprint arXiv:2408.00714  (2024)

\bibitem{rudenko2020human}
Rudenko, A., Palmieri, L., Herman, M., Kitani, K.M., Gavrila, D.M., Arras, K.O.: Human motion trajectory prediction: A survey. The International Journal of Robotics Research  \textbf{39}(8),  895--935 (2020)

\bibitem{ruegg2023bite}
R{\"u}egg, N., Tripathi, S., Schindler, K., Black, M.J., Zuffi, S.: Bite: Beyond priors for improved three-d dog pose estimation. In: Proceedings of the IEEE/CVF Conference on Computer Vision and Pattern Recognition. pp. 8867--8876 (2023)

\bibitem{salzmann2023robots}
Salzmann, T., Chiang, H.T.L., Ryll, M., Sadigh, D., Parada, C., Bewley, A.: Robots that can see: Leveraging human pose for trajectory prediction. IEEE Robotics and Automation Letters  \textbf{8}(11),  7090--7097 (2023)

\bibitem{salzmann2020trajectron++}
Salzmann, T., Ivanovic, B., Chakravarty, P., Pavone, M.: Trajectron++: Dynamically-feasible trajectory forecasting with heterogeneous data. In: European conference on computer vision. pp. 683--700. Springer (2020)

\bibitem{scholz2025plug}
Scholz, L.A., Mancienne, T., Stednitz, S.J., Scott, E.K., Lee, C.C.: Plug-and-play automated behavioral tracking of zebrafish larvae with deeplabcut and sleap: pre-trained networks and datasets of annotated poses. bioRxiv  (2025)

\bibitem{seff2023motionlm}
Seff, A., Cera, B., Chen, D., Ng, M., Zhou, A., Nayakanti, N., Refaat, K.S., Al-Rfou, R., Sapp, B.: Motionlm: Multi-agent motion forecasting as language modeling. In: Proceedings of the IEEE/CVF International Conference on Computer Vision. pp. 8579--8590 (2023)

\bibitem{simeoni2025dinov3}
Sim{\'e}oni, O., Vo, H.V., Seitzer, M., Baldassarre, F., Oquab, M., Jose, C., Khalidov, V., Szafraniec, M., Yi, S., Ramamonjisoa, M., et~al.: Dinov3. arXiv preprint arXiv:2508.10104  (2025)

\bibitem{songdenoising}
Song, J., Meng, C., Ermon, S.: Denoising diffusion implicit models. In: International Conference on Learning Representations (2021)

\bibitem{srivastava2015unsupervised}
Srivastava, N., Mansimov, E., Salakhudinov, R.: Unsupervised learning of video representations using lstms. In: International conference on machine learning. pp. 843--852. PMLR (2015)

\bibitem{sun2024ponymation}
Sun, K., Litvak, D., Zhang, Y., Li, H., Wu, J., Wu, S.: Ponymation: Learning articulated 3d animal motions from unlabeled online videos. In: European Conference on Computer Vision. pp. 100--119. Springer (2024)

\bibitem{thakkar2024adaptive}
Thakkar, N., Mangalam, K., Bajcsy, A., Malik, J.: Adaptive human trajectory prediction via latent corridors. In: European Conference on Computer Vision. pp. 297--314. Springer (2024)

\bibitem{tinbergen1963aims}
Tinbergen, N.: On aims and methods of ethology. Zeitschrift f{\"u}r tierpsychologie  \textbf{20}(4),  410--433 (1963)

\bibitem{tulyakov2018mocogan}
Tulyakov, S., Liu, M.Y., Yang, X., Kautz, J.: Mocogan: Decomposing motion and content for video generation. In: Proceedings of the IEEE conference on computer vision and pattern recognition. pp. 1526--1535 (2018)

\bibitem{vemula2018social}
Vemula, A., Muelling, K., Oh, J.: Social attention: Modeling attention in human crowds. In: 2018 IEEE international Conference on Robotics and Automation (ICRA). pp. 4601--4607. IEEE (2018)

\bibitem{von1953dancing}
Von~Frisch, K.: The dancing bees, vol.~354. A Harvest (1953)

\bibitem{walker2025}
Walker, J.C., Vélez, P., Cabrera, L.P., Zhou, G., Kabra, R., Doersch, C., Ovsjanikov, M., Carreira, J., Ginosar, S.: Generalist forecasting with frozen video models via latent diffusion  (2025), \url{https://arxiv.org/abs/2507.13942}

\bibitem{wang2025vggt}
Wang, J., Chen, M., Karaev, N., Vedaldi, A., Rupprecht, C., Novotny, D.: Vggt: Visual geometry grounded transformer. In: Proceedings of the IEEE/CVF Conference on Computer Vision and Pattern Recognition (2025)

\bibitem{wang2020imaginator}
Wang, Y., Bilinski, P., Bremond, F., Dantcheva, A.: Imaginator: Conditional spatio-temporal gan for video generation. In: Proceedings of the IEEE/CVF winter conference on applications of computer vision. pp. 1160--1169 (2020)

\bibitem{wen2023anypoint}
Wen, C., Lin, X., So, J., Chen, K., Dou, Q., Gao, Y., Abbeel, P.: Any-point trajectory modeling for policy learning (2023)

\bibitem{wu2023magicpony}
Wu, S., Li, R., Jakab, T., Rupprecht, C., Vedaldi, A.: Magicpony: Learning articulated 3d animals in the wild. In: Proceedings of the IEEE/CVF Conference on Computer Vision and Pattern Recognition. pp. 8792--8802 (2023)

\bibitem{xing2024aid}
Xing, Z., Dai, Q., Weng, Z., Wu, Z., Jiang, Y.G.: Aid: Adapting image2video diffusion models for instruction-guided video prediction. arXiv preprint arXiv:2406.06465  (2024)

\bibitem{xu2024flow}
Xu, M., Xu, Z., Xu, Y., Chi, C., Wetzstein, G., Veloso, M., Song, S.: Flow as the cross-domain manipulation interface. arXiv preprint arXiv:2407.15208  (2024)

\bibitem{yang2025tra}
Yang, J., Zhu, H., Wang, Y., Wu, G., He, T., Wang, L.: Tra-moe: Learning trajectory prediction model from multiple domains for adaptive policy conditioning. In: Proceedings of the IEEE/CVF Conference on Computer Vision and Pattern Recognition. pp. 6960--6970 (2025)

\bibitem{yang2023video}
Yang, S., Zhang, L., Liu, Y., Jiang, Z., He, Y.: Video diffusion models with local-global context guidance. arXiv preprint arXiv:2306.02562  (2023)

\bibitem{ye2024superanimal}
Ye, S., Filippova, A., Lauer, J., Schneider, S., Vidal, M., Qiu, T., Mathis, A., Mathis, M.W.: Superanimal pretrained pose estimation models for behavioral analysis. Nature communications  \textbf{15}(1), ~5165 (2024)

\bibitem{ye2024stdiff}
Ye, X., Bilodeau, G.A.: Stdiff: Spatio-temporal diffusion for continuous stochastic video prediction. In: Proceedings of the AAAI Conference on Artificial Intelligence. vol.~38, pp. 6666--6674 (2024)

\bibitem{yuan2024general}
Yuan, C., Wen, C., Zhang, T., Gao, Y.: General flow as foundation affordance for scalable robot learning. arXiv preprint arXiv:2401.11439  (2024)

\bibitem{zhang2025d4rt}
Zhang, C., Le~Moing, G., Koppula, S., Rocco, I., Momeni, L., Xie, J., Sun, S., Sukthankar, R., Barral, J.K., Hadsell, R., Ghahramani, Z., Zisserman, A., Zhang, J., Sajjadi, M.S.M.: Efficiently reconstructing dynamic scenes one d4rt at a time. arXiv preprint  (2025)

\bibitem{zholus2025tapnext}
Zholus, A., Doersch, C., Yang, Y., Koppula, S., Patraucean, V., He, X.O., Rocco, I., Sajjadi, M.S., Chandar, S., Goroshin, R.: Tapnext: Tracking any point (tap) as next token prediction. In: Proceedings of the IEEE/CVF International Conference on Computer Vision. pp. 9693--9703 (2025)

\bibitem{zuffi2019three}
Zuffi, S., Kanazawa, A., Berger-Wolf, T., Black, M.J.: Three-d safari: Learning to estimate zebra pose, shape, and texture from images" in the wild". In: Proceedings of the IEEE/CVF International Conference on Computer Vision. pp. 5359--5368 (2019)

\bibitem{zuffi2018lions}
Zuffi, S., Kanazawa, A., Black, M.J.: Lions and tigers and bears: Capturing non-rigid, 3d, articulated shape from images. In: Proceedings of the IEEE conference on Computer Vision and Pattern Recognition. pp. 3955--3963 (2018)

\bibitem{zuffi20173d}
Zuffi, S., Kanazawa, A., Jacobs, D.W., Black, M.J.: 3d menagerie: Modeling the 3d shape and pose of animals. In: Proceedings of the IEEE conference on computer vision and pattern recognition. pp. 6365--6373 (2017)

\bibitem{Zuffi_2024_CVPR}
Zuffi, S., Mellbin, Y., Li, C., Hoeschle, M., Kjellstr\"om, H., Polikovsky, S., Hernlund, E., Black, M.J.: Varen: Very accurate and realistic equine network. In: Proceedings of the IEEE/CVF Conference on Computer Vision and Pattern Recognition (CVPR). pp. 5374--5383 (June 2024)

\end{thebibliography}

\pagebreak
\begin{center}
    \LARGE \textbf{Appendix}
\end{center}

We provide an overview of data processing, motion distribution results, implementation details for our method and metrics, and additional qualitative and quantitative results. See https://motion-forecasting.github.io/ for video results.

\section{Data Processing details}
\label{sec:supp_data_processing}

Here we present an in-depth overview of our data processing pipeline, resulting in the MammalMotion dataset.

\subsection{Data Filtering}

Our pipeline begins with an initial quality filtering stage applied to the full (untrimmed) 539-hour MammalNet~\cite{chen2023mammalnet} dataset. Videos were excluded if they did not meet our minimum requirements for temporal and spatial resolution: a frame rate of at least $29.9$ FPS and a total resolution of $200,000$ pixels. 

We also remove all videos with a low dynamic range. The dynamic range of each video is computed by analyzing the pixel intensity distribution across all frames. For each frame, we first convert the image to grayscale and then calculate the dynamic range ratio using percentile-based thresholds to account for potential outliers. The dynamic range ratio $R$ for a frame is defined as:
$R = \frac{P_{99} - P_{1}}{I_{max} - I_{min}}$,
where $P_{99}$ and $P_{1}$ are the 99th and 1st percentiles of the pixel intensity distribution respectively, and $I_{max}$ and $I_{min}$ are the theoretical maximum and minimum intensity values possible for the image's data type. The final dynamic range measure for a video is computed as the mean of the frame-wise ratios. This metric provides a normalized measure between 0 and 1, where values closer to 1 indicate a wider effective dynamic range in the video content. We removed videos with a dynamic range value below $0.55$. 

After filtering according to the above criteria, we were left with just under 280 hours of video data.

\subsection{Shot Detection via Point Tracking}
After filtering at the video level, we divided the remaining videos into shots. Seeing that popular open-source libraries such as PySceneDetect fail to detect accurate shot boundaries on the difficult animal data, we developed a novel method for detecting shots based on the same point tracker that we used for obtaining point track training data.

Our algorithm works as follows: We use point-tracking to identify  temporal discontinuities in video sequences that indicate shot boundaries. Our algorithm operates by greedily dividing input videos into contiguous segments of up to 100 frames and systematically analyzing the temporal consistency of sparse point correspondences within each segment. For each video segment, the system samples 50 random query points at the first frame. These query points are then tracked forward in time using BootsTAPIR, which outputs point trajectories for the whole segment as well as visibility booleans. 

The shot change detection criterion is based on monitoring the percentage of visible points across all frames within each segment—when the visibility percentage drops below $6\%$ (less than 3 points are able to be tracked) for any frame, the algorithm identifies this as a shot change boundary, under the assumption that abrupt scene transitions cause widespread tracking failures due to the disappearance or significant transformation of visual features. When a segment contains multiple frames below the visibility threshold, only the earliest is recorded as the boundary. The segment window then restarts at that boundary frame $t'$, where new query points are sampled, allowing subsequent boundaries to be discovered in successive passes without any post-hoc merging. When no boundary is detected, the window advances by 100 frames, ensuring complete, gap-free coverage. 

Using this algorithm for shot detection has the added advantage that shots returned are ones where we will be able to track points. 

\subsection{Detection and Segmentation}

We now get a segmentation of every animal within each shot. Our pipeline begins with an initial animal detection stage for each video shot. We employ Grounding-DINO~\cite{liu2023grounding} on every frame, using the text prompt “animal” and a confidence threshold of $0.35$. Any shots without a single successful detection are discarded from the dataset.

We next identify frames within each shot that can be used to initialize a video segmenter on every animal in the shot.
To ensure tractability, shots longer than 1000 frames are first partitioned into $1000$-frame segments. 
We then developed a multi-stage heuristic to identify a frame where all animals are clearly visible and spatially distinct.

We first estimate the number of animals in the shot, $N$, by averaging the number of detections across all frames and rounding to the nearest integer. We then form a candidate pool of all frames containing exactly $N$ detections. From this pool, we isolate the top $10\%$ of frames with the lowest average Intersection over Union (IoU) among their bounding boxes. This step prioritizes frames where the animals exhibit minimal overlap. From this refined subset, we select the single frame with the highest mean detection confidence to serve as the definitive query frame.

Finally, we initialize VideoSAM~\cite{ravi2024sam} with the bounding boxes from the selected query frame. The resulting segmentation masks are then propagated bi-directionally to cover the entire shot.

\subsection{Point Tracking}

Once we have shots with animals segmented and tracked, we can track points within each animal. As point trackers are somewhat unreliable over long timeframes, we break each shot into-sub-shots of length up to 8 seconds(240 frames).  For each animal segmentation mask, we sample $500$ points across each sub-shot. To sample each point, we first sample uniformly in time (random frame indices within the shot). Then, we sample from the mask.

Our sampling strategy constrains query points to lie within animal segmentation masks and employs a distance transform-based weighting scheme to allow for sampling of thinner structures such as legs, tails, and heads. Specifically, $75\%$ of points are drawn according to an inverse distance transform distribution. Let $D(\mathbf{p})$ denote the Euclidean distance transform, i.e.\ the distance from pixel $\mathbf{p} \in M$ to the nearest boundary of segmentation mask $M$. The sampling probability is:
$P(\mathbf{p}) = \frac{1 / \bigl(D(\mathbf{p}) + \epsilon\bigr)}{\sum_{\mathbf{q} \in M} 1 / \bigl(D(\mathbf{q}) + \epsilon\bigr)},$
where $\epsilon = 10^{-6}$ ensures numerical stability. This assigns higher probability to pixels closer to mask boundaries, encouraging coverage of thin structures. The remaining $25\%$ of points are sampled uniformly within the mask to ensure coverage of interior regions.

Once query points are sampled, we track across the shot (up to 8 seconds) using BootsTAPIR~\cite{doersch2024bootstap}.

\subsection{Camera Stabilization}

While the tracked points are faithful to the animal pixels, the motion of the tracked points in pixel space confounds the motion of animals and the camera. Therefore, we employ a stabilization algorithm to disentangle the animal and camera motion, and train models on "stabilized" point tracks that only reflect the motion of animals.

Our approach first samples approximately $300$ background points from regions outside dilated animal segmentation masks, applying a $32$-pixel dilation buffer to ensure adequate separation from foreground motion. These background query points are evenly distributed across video frames and tracked using BootsTAPIR to establish correspondence across the temporal sequence. The resulting background point trajectories are then used to estimate inter-frame camera transformations through a robust RANSAC-based optimization process using publicly available code~\cite{doersch2024bootstap} that estimates a full homography (8 degrees of freedom). The camera motion estimation employs a reference frame approach where transformations are computed relative to a canonical middle frame, with iterative refinement passes to improve accuracy. To ensure high-quality transformations, we require a $>50\%$ average inlier ratio, and for the transformation matrix to be well-conditioned. We fail to stabilize $~7\%$ of the data and discard this before training. 

Once the homographies for each frame in a shot relative to a reference frame are computed, we can stabilize the point tracks at each timestep relative to the start of a time horizon. This enables us to understand how an animal moves, irrespective of camera motion. Please refer to the supplementary video for more results of our data processing pipeline, including point tracks before and after camera stabilization.

\section{Data Distribution}

\subsection{Motion Distribution}
\begin{figure*}[t]
  \centering
    \includegraphics[width=0.99\linewidth]{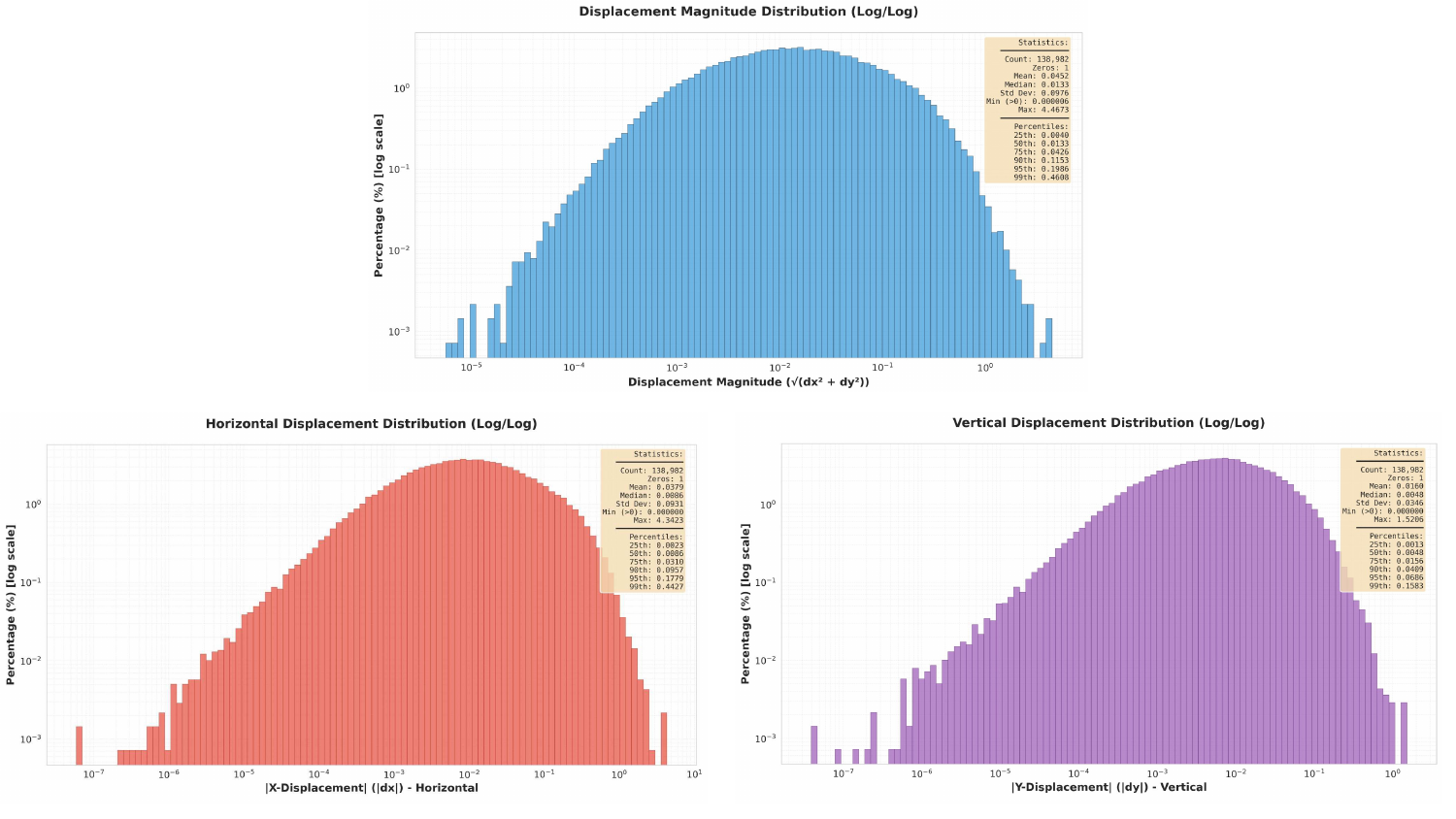}
  \caption{\textbf{Motion distribution in log-log.} We plot a histogram of animal displacement. The vertical axis is log frequency for all three plots. The horizontal axis is binned log displacement magnitude (top), binned log x-displacement (bottom left), and binned log y-displacement (bottom right).}
  \label{fig:motion_dist_xy}
  \vspace{-0.9em}
\end{figure*}

See Fig.~\ref{fig:motion_dist_xy} for our motion distribution for $x$, $y$, and overall animal motion. We see that for horizontal displacement, vertical displacement, and overall displacement, the distribution does not appear to follow a power law but rather a log-normal distribution.  We plot $x$ and $y$ displacement magnitudes separately to demonstrate that the log-normal distribution is not simply an artifact of the Euclidean displacement magnitude computation.

\section{Implementation details}

\subsection{Method Implementation Details}
\label{sec:method_impl_details}

\textbf{Architecture} Our model uses a DiT-B configuration with 12 transformer blocks, hidden dimension of 768, and 12 attention heads. $\sigma_v = 12.0, \sigma_o = 0.1$.

\textbf{Training.} We train with Adam optimizer with learning rate $5 \times 10^{-4}$, cosine annealing schedule with 5-epoch warmup, and batch size 64 distributed across 16 GPUs. We apply gradient clipping with norm 5.0. Training runs for 140 epochs.

\textbf{Exponential Moving Average (EMA).} We maintain an exponential moving average of model parameters with decay $\gamma = 0.9997$, a standard technique in diffusion models that stabilizes sample quality:
\begin{equation}
\theta_{\text{EMA}} \leftarrow \gamma \theta_{\text{EMA}} + (1 - \gamma) \theta.
\end{equation}
The EMA weights are used for all evaluation and inference.

\textbf{Data representation.} Each training example consists of $N = 320$ point tracks over a trajectory horizon of $T = 32$ timesteps (sampled at 15 FPS), conditioned on the first $T_c = 4$ timesteps. Tracks longer than $T$ timesteps are subsampled with stride 8. Tracks are normalized to $[0, 1]$ image coordinates within the animal's bounding box and stabilized via homography transformation. We handle variable numbers of valid points ($32 \leq N_{\text{valid}} \leq 320$) using attention masking to ignore padded points.

\subsection{Metric Implementation Details}

\paragraph{Fréchet Video Motion Distance (FVMD)}
We employ FVMD~\cite{liu2024fr} to evaluate motion consistency. Unlike image-level metrics (e.g., FVD), FVMD operates on explicit motion trajectories. For a set of point tracks, we compute velocity vectors $\mathbf{v}_t$ and acceleration vectors $\mathbf{a}_t = \mathbf{v}_{t+1} - \mathbf{v}_t$ in pixel space. 

Following the ``Dense 1D Histogram'' variant in~\cite{liu2024fr}, we discretize these vectors into histograms where bins correspond to motion directions (quantized into 8 angular bins). To emphasize significant movements, the contribution of each vector is weighted by its log-scaled magnitude. Specifically, for a magnitude $m_i = \|\mathbf{v}_i\|_2$, the weight $w_i$ is given by:
\begin{equation}
    w_i = \lceil \log_2(\min(m_i, 255) + 1) \rceil
\end{equation}
FVMD aggregates a separate histogram for each grid cell within a spatiotemporal grid with a $16\times16\times 8$ resolution, where $8$ corresponds to time. The final metric is defined as the Fréchet distance between the multivariate Gaussian approximations of the generated and ground truth motion feature distributions.

\section{Additional Results}

\subsection{Quantitative results}

We see results on our full dataset including medium and low motion buckets in table~\ref{tab:results_all_data_dist} (distribution level metrics) and~\ref{tab:results_all_data} (example level metrics), and results on the Panthera data subset for these motion buckets in~\ref{tab:results_panthera_dist_all_motion_buckets} and~\ref{tab:results_panthera_all_motion_buckets}.

\subsection{Qualitative results webpage}

We provide a \href{https://motion-forecasting.github.io/}{webpage} to further demonstrate the qualitative performance of our method. Each example first displays the ground truth video segment used to extract point tracks via our data processing pipeline, followed by our model’s predicted motions. The four conditioning timesteps (sampled at 15 FPS) are indicated by a grey border, while all subsequent frames are model predictions. Points predicted as occluded by our method are not rendered.

We showcase the following results:

1) \textbf{Diverse Species and Behaviors:} Results across a wide range of behaviors—including walking, mating, eating, fighting, and grooming—and across various species. Notably, our model demonstrates robust performance on rare species that are significantly underrepresented in the training set (e.g., fossa at $0.14\%$, tapir at $0.22\%$, and the caribou and eskimo dog at $0.025\%$). For context, even the most frequent species in our dataset (squirrel, giraffe, elephant, hamster, and deer) each comprise only approximately $3\%$ of the total data.

2) \textbf{Stochastic Motion Generation:} By varying the random seed while keeping the input image and motion history fixed, we demonstrate the model’s ability to generate diverse, physically plausible motion trajectories from the same initial context.

3) \textbf{Controllable Generation via Displacement Vectors}: We illustrate the model’s responsiveness to an optional 2D displacement vector. All results before these were generated without this prompting. Each set of results holds the input and random seed constant, but uses a different 2D displacement vector. The displacement vectors used, where $d = [d_x, d_y]$ is the ground truth displacement, are, from left to right, $d$, $-d$, $\frac{d}{2}$, and $2d$.

4) \textbf{Out-of-distribution generalization}: We evaluate our model’s zero-shot capabilities by prompting it with non-mammal animals, humans, and other objects. 

5) \textbf{Baseline Comparisons}: We provide side-by-side visualizations against the "Oracle Velocity" (our strongest non-learned baseline) and Track2Act trained on our full dataset. Comparisons with Track2Act use identical random seeds and motion history. Note that Track2Act and oracle velocity cannot handle occlusions, so all points are treated as visible.

6) \textbf{Comparison with Stable Video Diffusion~\cite{blattmann2023stable}}: While SVD produces high-quality results for common species (e.g., horses), it often struggles with rare species, frequently "shape-shifting" them into more common animals or failing to capture realistic behavioral patterns. We highlight these failure modes in species such as the hare ($0.39\%$ in our training dataset), elk ($1.2\%$), bison ($0.89\%$), and black rhino ($0.20\%$). We specifically use the Stable Diffusion XL model available through the interface available at https://stablediffusionweb.com/.

7) \textbf{Data Preprocessing and Camera Stabilization}: We visualize results from our data preprocessing pipeline, showcasing both raw outputs and results after camera stabilization. We observe that while many animals are detected, some are missed; furthermore, while the segmentation masks from VideoSAM are highly accurate, they are not perfect on this challenging data. Crucially, the camera stabilization of point tracks allows us to effectively disentangle animal motion from camera motion.

\begin{table}[t]
\centering
\caption{Quantitative results on \textbf{All Data}, distribution level. FD values are multiplied by $10^3$; Variance values are multiplied by $10^5$; FVMD values are divided by $10^3$. Best results in \textbf{bold}, second best \underline{underlined}. $\uparrow$ indicates higher is better; $\downarrow$ indicates lower is better.}
\label{tab:results_all_data_dist}
\small
\begin{tabular}{@{}ll ccccc@{}}
\toprule
\textbf{Selection} & \textbf{Method} & \textbf{FD (V)} $\downarrow$ & \textbf{FD (A)} $\downarrow$ & \textbf{Var (V)} & \textbf{Var (A)} & \textbf{FVMD} $\downarrow$ \\
\midrule
\multirow{7}{*}{High motion}
    & GT & - & - & 31.5 & 8.94 & - \\
    & No motion & 27.1 & 7.51 & 0 & 0 & 481.99 \\
    & Constant vel & 13.7 & 7.51 & 23.8 & 0 & 210.47 \\
    & WHN & 25.2 & \textbf{3.19} & 1.1 & 3.34 & 280.77 \\
    & Track2Act & \underline{11.8} & 4.81 & 9.89 & 1.63 & \underline{76.28} \\
    & Ours (uncond) & \textbf{8.96} & \underline{3.74} & 13.1 & 1.68 & \textbf{49.30} \\
    \cmidrule(lr){2-7}
    & Oracle vel & 12.1 & 7.51 & 19.8 & 0 & 326.80 \\
    & Ours (cond) & \textbf{4.86} & \textbf{3.33} & 28.3 & 2.14 & \textbf{40.24} \\
\cmidrule(lr){1-7}
\multirow{7}{*}{Medium motion}
    & GT & - & - & 1.32 & 1.07 & - \\
    & No motion & 1.14 & 0.897 & 0 & 0 & 139.91 \\
    & Constant vel & 1.43 & 0.897 & 1.03 & 0 & 89.23 \\
    & WHN & 0.559 & 0.679 & 1.21 & 3.65 & 33.86 \\
    & Track2Act & \underline{0.298} & \underline{0.325} & 0.592 & 0.319 & \underline{16.59} \\
    & Ours (uncond) & \textbf{0.257} & \textbf{0.298} & 0.396 & 0.251 & \textbf{12.90} \\
    \cmidrule(lr){2-7}
    & Oracle vel & 1.03 & 0.897 & 0.0825 & 0 & 163.67 \\
    & Ours (cond) & \textbf{0.197} & \textbf{0.28} & 0.613 & 0.314 & \textbf{12.13} \\
\cmidrule(lr){1-7}
\multirow{7}{*}{Low motion}
    & GT & - & - & 0.111 & 0.157 & - \\
    & No motion & 0.0957 & 0.132 & 0 & 0 & 80.13 \\
    & Constant vel & 0.124 & 0.132 & 0.0891 & 0 & 34.31 \\
    & WHN & 0.46 & 1.39 & 1.29 & 3.89 & 16.68 \\
    & Track2Act & \underline{0.0341} & \underline{0.0448} & 0.146 & 0.146 & \underline{7.73} \\
    & Ours (uncond) & \textbf{0.016} & \textbf{0.0309} & 0.0382 & 0.0365 & \textbf{4.11} \\
    \cmidrule(lr){2-7}
    & Oracle vel & 0.0886 & 0.132 & 0.00404 & 0 & 212.13 \\
    & Ours (cond) & \textbf{0.0148} & \textbf{0.0304} & 0.0416 & 0.0383 & \textbf{4.51} \\
\cmidrule(lr){1-7}
\multirow{7}{*}{Combined}
    & GT & - & - & 5.41 & 1.82 & - \\
    & No motion & 4.66 & 1.53 & 0 & 0 & 204.14 \\
    & Constant vel & \underline{2.59} & 1.53 & 4.11 & 0 & 89.77 \\
    & WHN & 5.34 & \textbf{0.691} & 1.23 & 3.7 & 94.7 \\
    & Track2Act & \underline{2.52} & 1.09 & 2.37 & 0.516 & \underline{26.17} \\
    & Ours (uncond) & \textbf{1.96} & \underline{0.877} & 2.94 & 0.453 & \textbf{17.0} \\
    \cmidrule(lr){2-7}
    & Oracle vel & 2.26 & 1.53 & 3.15 & 0 & 185.62 \\
    & Ours (cond) & \textbf{1.07} & \textbf{0.778} & 6.26 & 0.57 & \textbf{14.38} \\
\bottomrule
\end{tabular}
\end{table}

\begin{table}[t]
\centering
\caption{Quantitative evaluation on \textbf{All Data}, example-level metrics. Best results in \textbf{bold}, second best \underline{underlined}. For non-learned baselines, we report single-sample metrics; for WHN, and ours we report best of $K=5$.}
\label{tab:results_all_data}
\resizebox{\textwidth}{!}{%
\begin{tabular}{l l cccc}
\toprule
\textbf{Selection} & \textbf{Method} & \textbf{ADE} $\downarrow$ & \textbf{FDE} $\downarrow$ & \textbf{VMD} $\downarrow$ & \textbf{Avg PWT} $\uparrow$ \\
\midrule
\multirow{7}{*}{High motion}
    & No motion & 0.325 & 0.596 & 6.50 & 12.44\% \\
    & Constant vel & 0.286 & 0.591 & 5.02 & 11.94\% \\
    & WHN & 0.262 & 0.538 & 5.74 & 11.62\% \\
    & Track2Act & \underline{0.136} & \underline{0.294} & \underline{4.50} & \underline{21.79\%} \\
    & Ours (uncond) & \textbf{0.119} & \textbf{0.275} & \textbf{4.33} & \textbf{26.01\%} \\
    \cmidrule(lr){2-6}
    & Oracle vel & 0.110 & 0.156 & 7.04 & 14.70\% \\
    & Ours (cond) & \textbf{0.068} & \textbf{0.103} & \textbf{4.25} & \textbf{31.50\%} \\
\cmidrule(lr){1-6}
\multirow{7}{*}{Medium motion}
    & No motion & 0.032 & 0.057 & 5.29 & 48.53\% \\
    & Constant vel & 0.068 & 0.142 & 5.70 & 33.28\% \\
    & WHN & 0.035 & 0.049 & 4.75 & 34.49\% \\
    & Track2Act & \underline{0.023} & \underline{0.039} & \underline{3.69} & \underline{56.32\%} \\
    & Ours (uncond) & \textbf{0.020} & \textbf{0.035} & \textbf{3.57} & \textbf{59.45\%} \\
    \cmidrule(lr){2-6}
    & Oracle vel & 0.030 & 0.042 & 5.73 & 43.37\% \\
    & Ours (cond) & \textbf{0.016} & \textbf{0.021} & \textbf{3.51} & \textbf{63.05\%} \\
\cmidrule(lr){1-6}
\multirow{7}{*}{Low motion}
    & No motion & \underline{0.007} & \underline{0.011} & 3.44 & 83.75\% \\
    & Constant vel & 0.018 & 0.034 & 4.51 & 65.13\% \\
    & WHN & 0.023 & 0.024 & 4.19 & 41.93\% \\
    & Track2Act & \underline{0.006} & \underline{0.009} & \underline{2.48} & \underline{85.53\%} \\
    & Ours (uncond) & \textbf{0.005} & \textbf{0.008} & \textbf{2.27} & \textbf{88.48\%} \\
    \cmidrule(lr){2-6}
    & Oracle vel & 0.008 & 0.010 & 4.54 & 80.95\% \\
    & Ours (cond) & \textbf{0.004} & \textbf{0.006} & \textbf{2.26} & \textbf{90.19\%} \\
\cmidrule(lr){1-6}
\multirow{7}{*}{Combined}
    & No motion & 0.099 & 0.180 & 4.82 & 53.94\% \\
    & Constant vel & 0.104 & 0.215 & 5.02 & 41.15\% \\
    & WHN & 0.105 & 0.200 & 4.85 & 29.92\% \\
     & Track2Act & \underline{0.053} & \underline{0.110} & \underline{3.48} & \underline{56.59\%} \\
    & Ours (uncond) & \textbf{0.046} & \textbf{0.102} & \textbf{3.31} & \textbf{60.01\%} \\
    \cmidrule(lr){2-6}
    & Oracle vel & 0.042 & 0.058 & 5.57 & 51.74\% \\
    & Ours (cond) & \textbf{0.028} & \textbf{0.042} & \textbf{3.26} & \textbf{63.48\%} \\
\bottomrule
\end{tabular}%
}
\end{table}

\begin{minipage}{0.42\textwidth}

\end{minipage}

\begin{table}[t]
\centering
\caption{Quantitative results on \textbf{Panthera Data}, distribution level. FD values are multiplied by $10^3$; Variance values are multiplied by $10^5$; FVMD values are divided by $10^3$. Best results in \textbf{bold}, second best \underline{underlined}. $\uparrow$ indicates higher is better; $\downarrow$ indicates lower is better.}
\label{tab:results_panthera_dist_all_motion_buckets}
\small
\begin{tabular}{@{}ll ccccc@{}}
\toprule
\textbf{Selection} & \textbf{Method} & \textbf{FD (V)} $\downarrow$ & \textbf{FD (A)} $\downarrow$ & \textbf{Var (V)} & \textbf{Var (A)} & \textbf{FVMD} $\downarrow$ \\
\midrule
\multirow{9}{*}{High motion}
    & GT & - & - & 29.5 & 10.8 & - \\
    & No motion & 16.6 & 5.61 & 0 & 0 & 335.406 \\
    & Constant vel & 7.49 & 5.61 & 37.3 & 0 & 149.518 \\
    & WHN & 15.2 & \textbf{3.27} & 1.37 & 4.11 & 247.56 \\
    & ATM & 6.52 & 6.18 & 10 & 6.99 & 112.71 \\
    & Track2Act & \underline{6.32} & 5.06 & 8.01 & 0.446 & \underline{104.85} \\
    & Ours (uncond) & \textbf{3.71} & \underline{4.3} & 12.8 & 1.02 & \textbf{84.79} \\
    \cmidrule(lr){2-7}
    & Oracle vel & 5.7 & 5.61 & 20.9 & 0 & 218.73 \\
    & Ours (cond) & \textbf{2.82} & \textbf{4.19} & 16.6 & 1.18 & \textbf{79.38} \\
\cmidrule(lr){1-7}
\multirow{9}{*}{Medium motion}
    & GT & - & - & 1.26 & 0.931 & - \\
    & No motion & 0.681 & 0.484 & 0 & 0 & 91.93 \\
    & Constant vel & 0.822 & 0.484 & 0.959 & 0 & 54.51 \\
    & WHN & 0.726 & 1.16 & 1.45 & 4.35 & 46.47 \\
    & ATM & 0.494 & \textbf{0.384} & 0.28 & 0.475 & \underline{36.94} \\
    & Track2Act & \underline{0.421} & \underline{0.416} & 0.345 & 0.04 & 43.62 \\
    & Ours (uncond) & \textbf{0.405} & 0.417 & 0.179 & 0.027 & \textbf{26.85} \\
    \cmidrule(lr){2-7}
    & Oracle vel & 0.614 & 0.484 & 0.0708 & 0 & 107.49 \\
    & Ours (cond) & \textbf{0.389} & \textbf{0.414} & 0.184 & 0.0285 & \textbf{28.96} \\
\cmidrule(lr){1-7}
\multirow{9}{*}{Low motion}
    & GT & - & - & 0.142 & 0.173 & - \\
    & No motion & 0.077 & 0.09 & 0 & 0 & 46.0 \\
    & Constant vel & 0.0814 & 0.09 & 0.093 & 0 & \underline{15.20} \\
    & WHN & 0.527 & 1.56 & 1.44 & 4.33 & 19.89 \\
    & ATM & 0.0746 & 0.116 & 0.123 & 0.254 & 19.05 \\
    & Track2Act & \underline{0.0517} & \textbf{0.0731} & 0.0748 & 0.0294 & 24.36 \\
    & Ours (uncond) & \textbf{0.0444} & \underline{0.0776} & 0.026 & 0.00488 & \textbf{7.54} \\
    \cmidrule(lr){2-7}
    & Oracle vel & 0.0679 & 0.09 & 0.00643 & 0 & 119.12 \\
    & Ours (cond) & \textbf{0.0431} & \textbf{0.0774} & 0.0258 & 0.00499 & \textbf{9.95} \\
\cmidrule(lr){1-7}
\multirow{9}{*}{Combined}
    & GT & - & - & 6.93 & 2.66 & - \\
    & No motion & 3.77 & 1.38 & 0 & 0 & 149.53 \\
    & Constant vel & 1.86 & 1.38 & 8.58 & 0 & 62.51 \\
    & WHN & 3.37 & \underline{1.12} & 1.43 & 4.29 & 86.89 \\
    & ATM & 1.49 & 1.4 & 2.42 & 1.75 & \underline{35.50} \\
    & Track2Act & \underline{1.43} & 1.21 & 1.89 & 0.121 & 38.44 \\
    & Ours (uncond) & \textbf{0.874} & \textbf{1.05} & 2.94 & 0.226 & \textbf{24.82} \\
    \cmidrule(lr){2-7}
    & Oracle vel & 1.43 & 1.38 & 4.73 & 0 & 118.61 \\
    & Ours (cond) & \textbf{0.679} & \textbf{1.02} & 3.73 & 0.262  & \textbf{24.90} \\
\bottomrule
\end{tabular}
\end{table}

\begin{table}[t]
\centering
\caption{Quantitative evaluation on \textbf{Panthera}, example-level metrics. Best results in \textbf{bold}, second best \underline{underlined}. For non-learned baselines and ATM (single output), we report single-sample metrics; for WHN, Track2Act, and Ours we report best of $K=5$.}
\label{tab:results_panthera_all_motion_buckets}
\resizebox{\textwidth}{!}{%
\begin{tabular}{l l cccc}
\toprule
\textbf{Selection} & \textbf{Method} & \textbf{ADE} $\downarrow$ & \textbf{FDE} $\downarrow$ & \textbf{VMD} $\downarrow$ & \textbf{Avg PWT} $\uparrow$ \\
\midrule
\multirow{8}{*}{High motion}
    & No motion & 0.211 & 0.393 & 5.51 & 13.22\% \\
    & Constant vel & 0.193 & 0.413 & \underline{4.91} & \underline{16.73\%} \\
    & WHN & 0.215 & 0.393 & 5.82 & 10.24\% \\
    & ATM & 0.143 & 0.262 & 5.95 & 16.31\% \\
    & Track2Act & \underline{0.135} & \underline{0.245} & 5.04 & 16.72\% \\
    & Ours (uncond) & \textbf{0.107} & \textbf{0.209} & \textbf{4.77} & \textbf{20.68\%} \\
    \cmidrule(lr){2-6}
    & Oracle vel & 0.082 & \textbf{0.095} & 6.03 & 17.05\% \\
    & Ours (cond) & \textbf{0.067} & 0.097 & \textbf{4.61} & \textbf{27.31\%} \\
\cmidrule(lr){1-6}
\multirow{8}{*}{Medium motion}
    & No motion & \underline{0.022} & \underline{0.030} & 4.18 & \underline{58.72\%} \\
    & Constant vel & 0.044 & 0.080 & 4.78 & 44.33\% \\
    & WHN & 0.032 & 0.040 & 5.03 & 36.42\% \\
    & ATM & 0.025 & 0.037 & 4.51 & 51.24\% \\
    & Track2Act & 0.024 & 0.032 & \underline{3.99} & 53.69\% \\
    & Ours (uncond) & \textbf{0.020} & \textbf{0.027} & \textbf{3.82} & \textbf{60.91\%} \\
    \cmidrule(lr){2-6}
    & Oracle vel & 0.022 & 0.027 & 4.37 & 52.53\% \\
    & Ours (cond) & \textbf{0.016} & \textbf{0.019} & \textbf{3.75} & \textbf{63.66\%} \\
\cmidrule(lr){1-6}
\multirow{8}{*}{Low motion}
    & No motion & \underline{0.007} & \underline{0.010} & 2.71 & \underline{84.71\%} \\
    & Constant vel & 0.013 & 0.024 & 3.55 & 70.99\% \\
    & WHN & 0.022 & 0.023 & 4.45 & 42.40\% \\
    & ATM & 0.010 & 0.016 & 3.29 & 72.06\% \\
    & Track2Act & 0.008 & 0.012 & \underline{2.70} & 76.87\% \\
    & Ours (uncond) & \textbf{0.006} & \textbf{0.009} & \textbf{2.57} & \textbf{86.10\%} \\
    \cmidrule(lr){2-6}
    & Oracle vel & 0.007 & 0.009 & 3.40 & 82.43\% \\
    & Ours (cond) & \textbf{0.005} & \textbf{0.007} & \textbf{2.56} & \textbf{87.76\%} \\
\cmidrule(lr){1-6}
\multirow{8}{*}{Combined}
    & No motion & 0.076 & 0.138 & 4.02 & \underline{54.55\%} \\
    & Constant vel & 0.079 & 0.164 & 4.33 & 46.16\% \\
    & WHN & 0.086 & 0.146 & 5.05 & 30.43\% \\
    & ATM & 0.057 & 0.101 & 4.48 & 48.39\% \\
    & Track2Act & \underline{0.053} & \underline{0.092} & \underline{3.81} & 51.13\% \\
    & Ours (uncond) & \textbf{0.042} & \textbf{0.078} & \textbf{3.62} & \textbf{58.11\%} \\
    \cmidrule(lr){2-6}
    & Oracle vel & 0.035 & 0.042 & 4.51 & 53.14\% \\
    & Ours (cond) & \textbf{0.028} & \textbf{0.039} & \textbf{3.55} & \textbf{61.67\%} \\
\bottomrule
\end{tabular}%
}
\end{table}

\end{document}